\documentclass[runningheads]{llncs}

\usepackage[mobile]{eccv}

\usepackage{eccvabbrv}

\usepackage{graphicx}
\usepackage{booktabs}
\usepackage{multirow}
\usepackage{amsmath}%%新增
\usepackage{amssymb}%%新增
\usepackage{dsfont}%%新增
\usepackage{wrapfig}
\usepackage{makecell}
\usepackage{subcaption}
\usepackage{graphicx}
\usepackage{marvosym}
\usepackage{ragged2e}

\usepackage[accsupp]{axessibility}  % Improves PDF readability for those with disabilities.

\usepackage{hyperref}

\usepackage{orcidlink}

\begin{document}

\title{Denoising-Enhanced Coarse-to-Fine Infrared Small Target Detection with Attention Prior-Guided Knowledge Distillation} 

\titlerunning{ECFNet}

\author{Houzhang Fang\inst{1}\orcidlink{0000-0002-7949-8846} 
\and
Ruixuan Huang\inst{1}\textsuperscript{(\Letter)}\thanks{Corresponding author.}
\and
Qiuhuan Chen\inst{1} \and Xiaolin Wang\inst{1} 
\and Yi Chang\inst{2}\orcidlink{0000-0001-8542-5937} 
\and Luxin Yan\inst{2}\orcidlink{009-0009-6446-1715}
}

\authorrunning{H.~Fang et al.}

\institute{Xidian University, Xi’an, China\\
\email{houzhangfang@xidian.edu.cn, \{hrx, cqh, wxl\}@stu.xidian.edu.cn}
\and
Huazhong University of Science and Technology, Wuhan, China\\
\email{\{yichang, yanluxin\}@hust.edu.cn}
}

\maketitle

\begin{abstract}
Infrared small target detection (IRSTD) in high-resolution images is crucial for many practical applications, such as surveillance of unmanned aerial vehicles (UAVs) and UAV-based ground monitoring. However, IRSTD remains challenging due to the small size and weak features of targets, as well as significant interference from complex dynamic backgrounds. Existing detection methods often suffer from redundant computations on non-target background regions and insufficient exploitation of target context information, which limits their performance in complex backgrounds. To address these issues, we propose an efficient coarse-to-fine infrared small target detection framework with attention prior-guided knowledge distillation, termed ECFNet. In the coarse stage, we design a region binary classification network (RBCN) on grid-based multi-scale feature maps to efficiently recognize target-containing context region proposals. Moreover, we introduce a novel denoising-assisted training strategy that incorporates noisy ground-truth (GT) masks into RBCN feature maps and trains the network to reconstruct the original GT masks through a denoising task, thereby encouraging it to explicitly learn target-background context and thus better distinguish target proposals from background regions. In the fine stage, we customize a lightweight target detector to the coarse stage’s region proposals for balancing accuracy and efficiency. Furthermore, we propose a knowledge distillation strategy guided by the teacher-student cross-attention prior. This mechanism directs the student to focus on critical target regions, thereby enhancing the discriminative feature representation for infrared small targets. Extensive experiments on three real infrared datasets demonstrate that our method outperforms both existing single-stage and two-stage approaches while maintaining high real-time processing efficiency. \textbf{Code:} \url{https://github.com/IVPLabs/ECFNet}.
  \keywords{Infrared small target detection \and Coarse-to-fine framework \and Denoising-assisted training \and Prior-guided knowledge distillation}
\end{abstract}

\section{Introduction}
\label{sec:intro}

\begin{figure}[!t]
\centering
\includegraphics[width=3.3in,keepaspectratio]{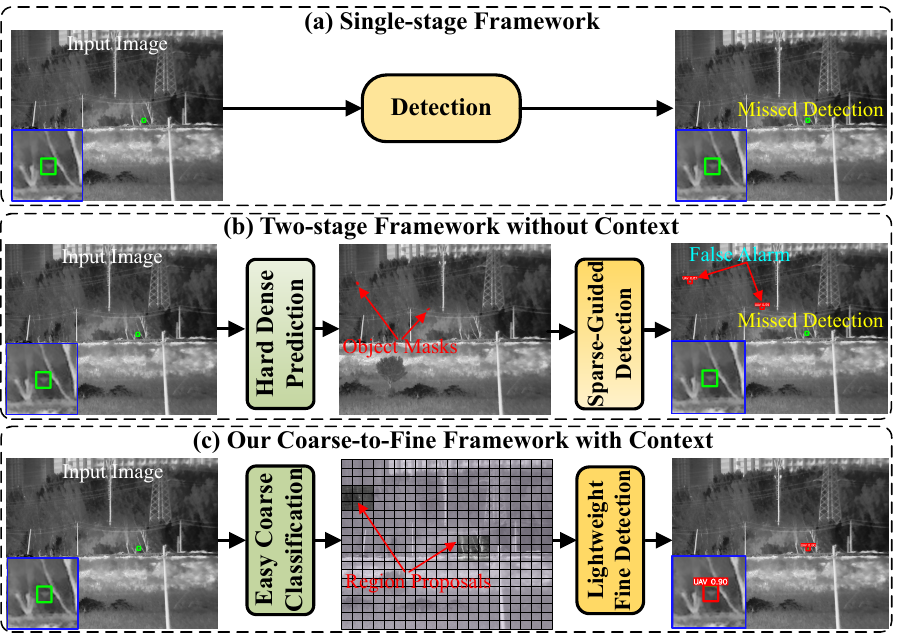} %\vspace{-0.6em}
\caption{Three main categories of methods for IRSTD. (a) Single-stage framework: detection models are directly applied to infrared images. (b) Two-stage framework without context: dense prediction (\textit{e.g.}, segmentation) produces a target mask, followed by sparse detection on the mask. (c) Coarse-to-fine framework with context: the coarse stage generates target region proposals with context, and the fine stage applies a lightweight detector on the proposals. %Previous methods overlook context in the coarse stage, 
Our approach leverages contextual information to enhance the semantic understanding of the proposals, thereby improving target proposal recognition and providing valuable proposals for the fine detection.
}
\label{summary}
\end{figure}

Infrared small target detection (IRSTD) in high-resolution images is an important perception technology in many applications, such as surveillance of unmanned aerial vehicles (UAVs) \cite{2022TIMFang,2023ACMMMFang,2023TIIFang,2025CVPRUniCD,2026AAAI_JFD3,2026AAAI_TDCNet}, UAV-based ground surveillance \cite{2025TIPESODLiu}, and remote sensing \cite{2024TGRSDTNet_Zhang,AAAI2025HS-FPNShi}. IRSTD remains highly challenging due to the small size and weak features of targets, as well as severe interference from complex dynamic backgrounds. These factors often result in both detection failures and false alarms \cite{2023ACMMMFang,2023TIIFang,AAAI2025HS-FPNShi}. 
%As the special imaging principle of infrared images, small targets often present spectral characteristics highly similar to the background and lack important texture information.

Advanced detection methods have been introduced to address this challenge. Current methods can be broadly categorized into single-stage \cite{2022TIMFang, 2023TIIFang,2024ECCVHuangDQ-DETR, 2023TIPXUIUNet, 2024TGRSyangEFLNet, 2022CVPRZhangISNet, 2025AAAIYangPinwheel,2023CVPRYingLESPS,2025CVPRUniCD} and two-stage approaches \cite{2025TIPESODLiu, 2022CVPRYangQueryDet, 2022TGRSWangIAAN}. Single-stage approaches focus on designing complex network architectures to extract infrared small target (IRST) features. However, IRSTs occupy only a small portion of the image, and thus most redundant feature computations are wasted on non-target background regions \cite{2025TIPESODLiu}. This not only limits detection accuracy but also hampers real-time performance, especially when processing high-resolution images (\cref{summary}(a)). 

Two-stage approaches typically first leverage dense prediction \cite{2023CVPRWMeethalCascaded, 2022TGRSWangIAAN, 2015NIPSRenFasterRCNN} (\textit{e.g.}, segmentation) to generate initial mask proposals, which are subsequently refined through lightweight detectors \cite{2025TIPESODLiu} (\cref{summary}(b)). However, dense prediction suffers from two primary limitations. First, during the initial stage, they rely predominantly on target-specific features for target proposal extraction, often neglecting crucial contextual cues. Considering the inherently subtle feature representation of small targets, this lack of context severely restricts the prediction precision. Second, while existing approaches \cite{2025ICCVAasishDM-EFS, 2025TIPESODLiu,2023ICCVXiangCFINet, 2022CVPRYangQueryDet} employ specialized network architectures of different complexities to boost performance, they lack an explicit mechanism to mitigate the severe interference from background clutter inherent in infrared images, inevitably leading to both false alarms and missed detections. Knowledge distillation (KD) is an effective strategy for enhancing feature representation in lightweight detectors. However, conventional KD methods typically focus on minimizing global feature discrepancies between teachers and students while overlooking the prior knowledge associated with small target regions, leading to limited performance gains.

%However, the former often leads to high computational cost, while the latter suffers from limited representational capacity, thereby impeding  efficiency or accuracy.  Existing sparse detection pipelines often face a dilemma: narrow cropping windows discard the discriminative context necessary for identifying dim targets, while expansive dense predicting introduces prohibitive computational overhead on vast non-target regions. 

% Furthermore, sparse detection in the second stage is susceptible to missed detections and false alarms due to the input of weak target features that lack sufficient context, which further compromises the detection accuracy.

To address the above issues, we propose an efficient coarse-to-fine detection framework (ECFNet) to effectively detect IRSTs (\cref{summary}(c)). Specifically, in the coarse stage, we construct a region binary classification network (RBCN) to reformulate the problem into patch-level binary recognition on non-overlapping grid region-based multi-scale feature maps, identifying context region proposals that contain the targets. This significantly reduces the number of predictions required compared to dense prediction \cite{2025TIPESODLiu, 2022CVPRYangQueryDet}. Recognizing that infrared small targets are often visually indistinguishable from complex backgrounds, we propose a denoising-assisted training (DAT) strategy. By injecting target-like noise to corrupt the ground-truth (GT) masks and integrating them into the RBCN feature space, DAT drives the network to reconstruct the original GT masks through a denoising task. This strategy encourages the network to explicitly model contextual target–background relationships, thereby improving its ability to discriminate target region proposals from complex backgrounds. It significantly improves the quality of region proposals and suppresses complex backgrounds. Moreover, this auxiliary denoising task stabilizes early optimization and effectively accelerates the convergence. To handle the issue of irregular targets being divided across multiple grid regions, we introduce ExpSlicer, which aggregates multi-scale pyramid region feature maps from the RBCN output to construct a complete target region. This approach ensures that the target is fully contained within a large single region patch while preserving critical contextual cues.

% To further improve the RBCN's capability of distinguishing target region proposals from complex backgrounds, we design a denoising-assisted training (DAT) strategy by introducing three additional auxiliary training branches. DAT first constructs noised masks by incorporating target-like artificial noises into the ground truth (GT) masks, and then performs a denoising task to reconstruct the GT masks. This auxiliary task leverages contextual information of the targets to enhance the semantic understanding of the proposals \cite{2024ECCVShuoQueryDenoise}. This auxiliary task leverages contextual information of the targets to enhance the semantic understanding of the proposals
%, and provides high-quality proposals for the fine detection

In the fine stage, we tailor a lightweight target detector to the coarse stage’s region proposals for reliable IRSTD. Unlike mask-based refinement \cite{2025TIPESODLiu, 2022TGRSWangIAAN}, our method leverages the complete target region during the fine stage, enhancing robustness to region recognition errors from the coarse stage. Additionally, our  lightweight detector processes only a small local region instead of the whole image, achieving more accurate and efficient target detection.

Since lightweight detectors typically suffer from limited representation capability, we propose an attention prior-guided knowledge distillation (APKD) strategy, leveraging cross-attention to propagate the teacher’s spatial attention priors over critical target regions to the student. 
%To further improve detection performance while maintaining high efficiency, we propose an attention prior-guided knowledge distillation (APKD) strategy. % to guide the lightweight student detector to focus on the critical target regions. %Most existing knowledge distillation (KD) methods for dense prediction tasks align the feature maps from the student and teacher networks, which often struggles to effectively enhance the student’s attention to target regions. 
%Given the teacher detector’s superior feature extraction capability, it is better equipped to capture the position, structure, and semantics of targets. 
In this work, we first calculate the global cross-attention between the teacher and student features to model their semantic dependencies, enabling the student to benefit from the teacher’s holistic understanding of the scene. We then modulate the student's features with the cross-attention weights, guiding the student detector to focus on the critical target regions emphasized by the teacher. The APKD between the teacher’s features and the modulated student features further enhances the student's accurate perception of key target regions. Experimental results on infrared UAV, Car and IRSTD-1k \cite{2022CVPRZhangISNet} datasets demonstrate that our framework ECFNet outperforms state-of-the-art (SOTA) single-stage and two-stage  methods while maintaining high detection efficiency.

Our contributions are summarized in three-fold:
\begin{itemize}
\item We propose a novel coarse-to-fine target detection framework that significantly enhances the detection performance of IRSTs under complex backgrounds while efficiently reducing computational costs, particularly in detecting high-resolution images.
\item We propose a novel denoising-assisted training strategy that reconstructs GT masks from noisy counterparts while explicitly modeling target–background context, improving discrimination in complex backgrounds and accelerating convergence for IRST detection.
\item We introduce a novel teacher-student cross-attention-guided KD mechanism that guides the student model to focus more effectively on critical target regions, thereby enhancing its ability to represent discriminative features of IRSTs.
\end{itemize}

\section{Related Work}
\subsection{Infrared Small Target Detection}
% In recent years, IRSTD has witnessed significant progress. However, IR images typically suffer from low signal-to-clutter ratios, with small targets often buried in complex backgrounds, making accurate detection highly challenging. Several one-stage methods have been proposed to address this. For instance, Yang et al. introduced PConv~\cite{2025AAAIYangPinwheel} to enhance feature extraction capabilities, and EFLNet proposed a Dynamic Head~\cite{2024TGRSyangEFLNet} for adaptive multi-scale feature fusion, improving spatial information representation and task correlation. MSHNet~\cite{2023CVPRLiuMSHNet} leverages a multi-scale head to improve target localization, while ISNet~\cite{2022CVPRZhangISNet} integrates shape reconstruction via edge and TOAA blocks within a U-Net framework.
% However, one-stage methods often waste considerable computational resources on irrelevant background regions. To address this, we propose a coarse-to-fine two-stage framework that reduces computational redundancy while ensuring high detection performance.

Recent years have seen significant progress in IRSTD, with many one-stage detectors proposed to enhance feature extraction and improve localization accuracy.  These methods typically process the entire image in a dense manner, aiming to capture target-related information directly from complex backgrounds~\cite{2025AAAIYangPinwheel, 2024CVPRLiuMSHNet,2022CVPRZhangISNet, 2024TGRSyangEFLNet,2025CVPRLi,2024TGRSChen}.  However, due to the low signal-to-clutter ratio and small target size in infrared images, one-stage detectors often incur substantial computational cost on irrelevant background regions.  To mitigate this inefficiency, we propose a coarse-to-fine two-stage framework that focuses computational resources on target-relevant regions while maintaining high detection performance.

\subsection{Two-Stage Detection Methods}
Recently, advanced two-stage detection architectures~\cite{2025ICCVAasishDM-EFS, 2023CVPRWMeethalCascaded, 2023ICCVXiangCFINet, 2022TGRSWangIAAN, 2015NIPSRenFasterRCNN} have been developed. ESOD~\cite{2025TIPESODLiu} employs a single depth-wise convolution~\cite{2017CVPRCholletDWConv} to perform dense predictions in the first stage, generating target masks as region proposals, followed by a sparse detection head in the second stage. However, the dense prediction is computationally expensive and challenging to optimize, while the sparse detection often discards critical contextual information, leading to increased missed detections and false alarms. QueryDet~\cite{2022CVPRYangQueryDet} dynamically selects the most appropriate feature level for each object through a query mechanism, enhancing overall small targets detection performance. While these methods prioritize either global search or local refinement, they often overlook the computational overhead of dense prediction ~\cite{2017NIPSVasAttention} or the context loss in sparse cropping.  Therefore, we construct an efficient coarse-to-fine architecture, using a lightweight detection model while retaining key contextual information. %~\cite{2016CVPRBellContext}

% But the global modeling based on Transformer~\cite{2017NIPSVasAttention} brings about a huge amount of computation.

\begin{figure*}[tb]
\centering
\includegraphics[width=\linewidth]{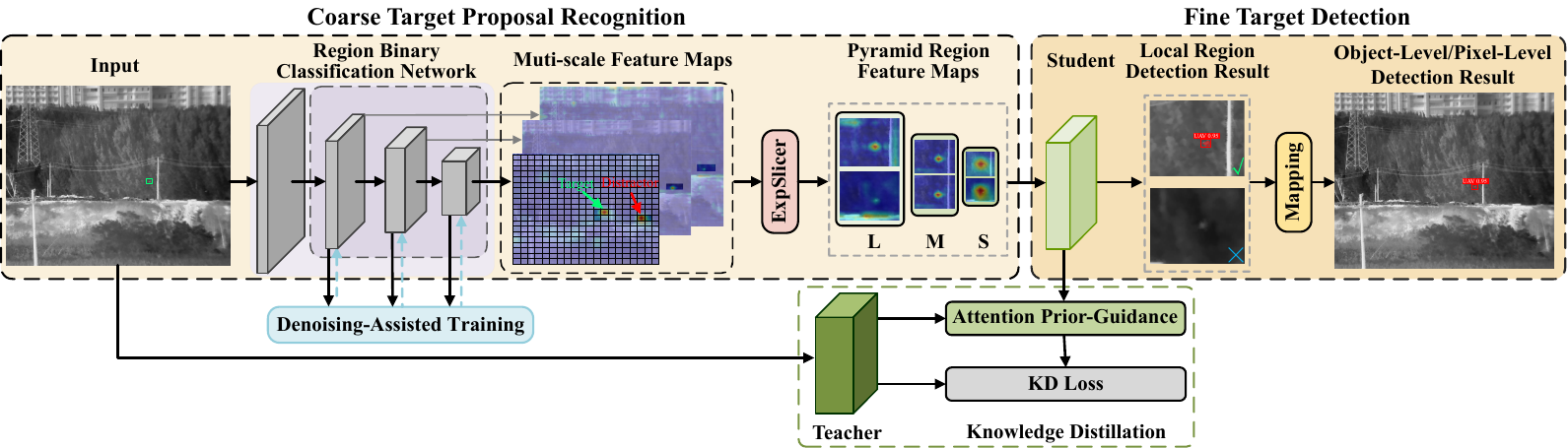} %\vspace{-2em}
\caption{The overall pipeline of ECFNet. ECFNet first employs the RBCN to generate context region proposals in the feature domain, followed by slicing these regions into square patches using the ExpSlicer. We leverage the DAT to further enhance the ability of RBCN to distinguish target proposals from complex backgrounds. In the fine stage, we customize a lightweight object-level or pixel-level detector in conjunction with an attention prior-guided knowledge distillation strategy to achieve more precise localization. Finally, we map the local region detection result back to the original image. }
\label{coarse_to_fine_framework}
\end{figure*}

\subsection{Knowledge Distillation}
In recent years, feature domain KD~\cite{2015ICLRAdrianaFeatureKD} has been exploited to improve the performance of lightweight student models by transferring the knowledge of the teacher model. DCSF~\cite{2025AAAIDaiDCSF} adopts a cross-layer feature fusion strategy to bridge multi-scale semantic differences. Fang et al.~\cite{2023TIIFang} proposed scale-specific KD to enhance multi-scale target feature representation by concatenating the student feature map and the teacher feature map after channel attention modulation. However, such conventional distillation strategies frequently overlook the intrinsic semantic misalignment and representational discrepancy between the teacher and student models.
Hence, we propose a dynamic knowledge distillation strategy based on cross-attention \cite{2025CVPRLouOverlock}, which explicitly models the semantic interaction between teacher and student features, enhancing target perception while suppressing background interference.

\section{Coarse-to-Fine Detection Method}
\subsection{Overall Architecture}

In this section, we propose ECFNet, an efficient coarse-to-fine infrared small target detection framework, as illustrated in ~\cref{coarse_to_fine_framework}. In the coarse stage, the RBCN distinguishes foreground regions from background, and the ExpSlicer selects high-confidence patches for refinement. In the fine stage, a lightweight IRSTD model performs precise localization on the sliced patches, whose results are then mapped back to the input image. Furthermore, the DAT and APKD modules are incorporated into the coarse and fine stages, respectively, to further enhance detection performance.

\subsection{Region Binary Classification Network (RBCN)}
\begin{figure}[tb]
\centering
\includegraphics[width=0.95\textwidth]{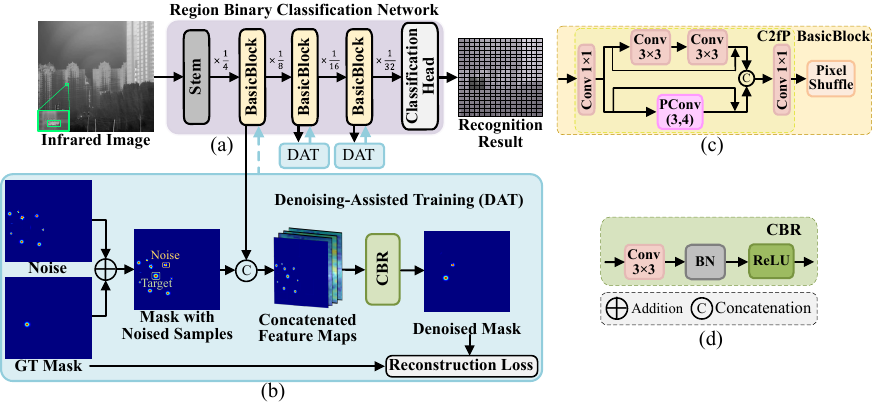} %\vspace{-1em}
\caption{Details of DAT. Multi-scale denoising prediction and GT are utilized for loss calculation. C2fP BasicBlock is a C2f-style module that mainly fuses a two-3$\times$3-conv residual path with a PConv residual path. CBR denotes the Conv–BN–ReLU block. PConv(3, 4) \cite{2025AAAIYangPinwheel} means the first PConv kernel is 3 and the second is 4.}
\label{DAT_fig}
\end{figure}

Conventional two-stage detection frameworks typically adopt a dense prediction step (\textit{e.g.}, semantic segmentation) followed by sparse convolution-based detection. However, pixel-level classification  introduces substantial computational overhead, making it difficult to balance accuracy and efficiency. Moreover, object masks lack essential contextual information. To overcome these limitations, we propose the RBCN, as illustrated in ~\cref{DAT_fig}(a), which reformulates pixel-wise prediction as a grid-level binary classification task, thereby simplifying the detection process and reducing computational cost. The backbone of RBCN adopts a four-stage hierarchical architecture. The spatial resolution is progressively reduced while the channel width increases across stages to capture higher-level semantic representations. In the second to the fourth stages, three BasicBlocks are employed to facilitate efficient feature propagation and improve representation capacity. Specifically, the input image is partitioned into \( N \times N \) non-overlapping grids, and RBCN outputs a probability mask \( M_{\text{pred}} \), where each value denotes the likelihood of a target existing within the corresponding grid. A binary coding strategy is then applied, assigning a value of 1 to grids exceeding a predefined threshold (we set 0.5 here) and 0 otherwise. Additionally, multi-scale feature maps are extracted for downstream processing. For the coarse stage, we design a joint loss function combining binary cross-entropy (BCE)~\cite{2025AAAIZhangMOCID} and IoU loss~\cite{2016ACMYuIoU}: $\mathcal{L}_{\text{coarse}} = \mathcal{L}_{\text{BCE}}({M}_{pred}, {M}_{GT}) + \lambda_1 \cdot \mathcal{L}_{\text{IoU}}({M}_{pred}, {M}_{GT}),$
where $\lambda_1$ denotes a learnable weight initialized to $1.0$ that balances the two terms.

% \begin{figure}[htbp]
%      \centering
%      % 第一张子图 (a)
%      \begin{subfigure}[c]{0.49\textwidth}
%          \centering
%          \includegraphics[width=\textwidth]{DAT_ab.pdf}
%      \end{subfigure}
%      \hspace{-0.5em} %较小的固定间距
%      % 第二张子图 (b)
%      \begin{subfigure}[c]{0.49\textwidth}
%          \centering
%          \includegraphics[width=\textwidth]{}
%      \end{subfigure}   
%      \caption{Details of DAT. Multi-scale denoising prediction and GT are utilized for loss calculation. C2fP BasicBlock is a C2f-style module that mainly fuses a two-3$\times$3-conv residual path with a PConv residual path. CBR denotes the Conv–BN–ReLU block. PConv(3, 4) \cite{2025AAAIYangPinwheel} means the first PConv kernel is 3 and the second is 4.}
%      \label{DAT_fig}
% \end{figure}

\subsection{Denoising-Assisted Training (DAT)}
IRSTs often exhibit high spectral similarity with background clutter, posing significant challenges for effective feature discrimination. To provide high-quality region proposals for the fine detection stage, we introduce a DAT module in the coarse stage, as illustrated in ~\cref{DAT_fig}(b).
To guide the model in suppressing target-like distractors, DAT synthesizes noise $N_{i}^{l}$ based on a 2D Gaussian function~\cite{2023TPAMIYangSCRDet}:
$N_{i}^{l}(x, y) = \exp\left(-\frac{(x - x_i)^2 + (y - y_i)^2}{2\sigma^2}\right)$,
where $(x_i, y_i)$ denotes the center of the $i$-th synthetic distractor and $\sigma$ controls the spatial spread. The superscript $l \in \{1,2,3\}$ corresponds to multi-scale feature levels produced by the RBCN.
These noises are added element-wise to the GT mask $M^{l}_{\text{GT}}$ and concatenated with RBCN features $F^{l}_{\text{RBCN}}$:
$F^{l}_{\text{DAT}} = \text{cat}\left(F^{l}_{\text{RBCN}},\ N^{l} + M^{l}_{\text{GT}} \right),$
where $\text{cat}(\cdot)$ denotes channel-wise concatenation. The obtained features are sent to the subsequent auxiliary branch to obtain the denoising mask $M^{l}_{\text{DAT}}$. The DAT module is supervised by resolution-aligned GT masks $\text{GT}^l$, with the reconstruction loss computed as: $\mathcal{L}_{\text{DAT}} = \sum_{l=1}^{3} \left( \mathcal{L}_{\text{BCE}}({M}^l_\text{DAT},{M}^l_{\text{GT}}) + \lambda_2 \cdot \mathcal{L}_{\text{IoU}}({M}^l_\text{DAT},{M}^{l}_{\text{GT}}) \right),$
where $\lambda_2$ is a learnable weighting coefficient that balances the reconstruction objectives. By injecting structured target-like noise into the feature space, DAT simulates realistic distractors. Since real and synthetic targets share similar local features, the network is forced to exploit surrounding context rather than isolated responses, thereby improving target-background discrimination.
% By injecting structured, target-like noise into the feature domain, DAT simulates real-world distractors (\textit{e.g.}, birds or clutter) and improves the model’s discriminative ability while suppressing background interference. 

% \noindent\textbf{Remark.} Unlike DN-DETR~\cite{2024TPAMILiDNDETR}, which generates noisy queries by duplicating ground-truth instances with perturbed labels and boxes, our method injects spatially correlated noise that mimics target feature distributions. Beyond the noise generation strategy, the objectives also differ: DN-DETR enhances object-query matching robustness, whereas our DAT strengthens the discriminability of region proposals against target-like background clutter.

\noindent\textbf{Discussion.} Unlike DN-DETR~\cite{2024TPAMILiDNDETR}, which introduces denoising queries for matching stabilization, DAT operates on region features by injecting target-like perturbations to simulate infrared clutter, thereby enhancing target-background discriminability under GT-mask supervision.

\subsection{Context-Preserving Region Expslicer}
\begin{figure}[tb]
\centering
\includegraphics[width=3.7in,keepaspectratio]{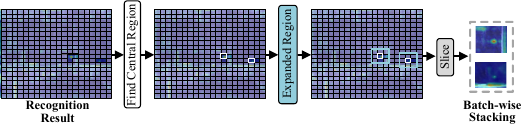} %\vspace{-1em}
\caption{ExpSlicer selects grids with the highest predicted target coverage (highlighted in white) from RBCN outputs as region centers and extracts $3\times3$ areas (shown in blue box) around them.}
\label{Expslicer}
\end{figure}

After obtaining the high-quality  region proposals $M_{pred}$, the customized ExpSlicer crops multi-scale feature maps $F^{l}_{RBCN}$ into fixed-size rectangular patches that preserve complete targets, as shown in ~\cref{Expslicer}.

Given the multi-scale feature maps \( F^{l}_{RBCN} \in \mathbb{R}^{B \times C \times H \times W} \) and \( M_{pred} \in \{0,1\}^{B \times 1 \times N \times N} \) generated by RBCN, we slice patches of feature maps centered on the positive positions of \( M_{pred} \) based on target coverage. $B, C, H$, and $W$ denote the batch size, channels, height, and width, respectively. Each spatial patch in \( M_{pred} \) corresponds to a size of \( h_p \times w_p \) in \( F^{l} \), such that: $
h_p = {H}/{N}, w_p = {W}/{N}.$
For every valid position \( (g_x, g_y) \) such that \( M_{pred}[b, 0, g_x, g_y]=1 \), we compute its center in the feature map as:
$(x_c, y_c) = (g_x \cdot w_p, \ g_y \cdot h_p)$. Then, a region of size \( (k \cdot w_p) \times (k \cdot h_p) \), centered at \((x_c, y_c)\), is extracted, where \(k\) denotes the side length of the patch centered at \((x_c, y_c)\). The hyperparameter \(k\) is set to 3 and \(N\) is set to 20 in our experiments.
 The cropped areas will be marked to prevent repeated cropping in subsequent operations. Let \( A \) be the total number of activated positions in $M_{pred}$. All patches in the same layer \( \{P_i\}_{i=1}^A \in \mathbb{R}^{C \times (k \cdot h_p) \times (k \cdot w_p)} \) are stacked along the batch dimension to construct an aggregated representation $F^{l}_{\text{agg}}\in \mathbb{R}^{(A\cdot C) \times (k \cdot h_p) \times (k \cdot w_p)}$. Global non-maximum suppression eliminates redundant predictions arising from overlapping regions and closely targets. For boundary targets, we use zero-padding to ensure the same size of the patches. 

\noindent\textbf{Discussion.} Such slicing strategy has two advantages. First, it ensures that the target is completely fed into the subsequent lightweight detection model without affecting the detection accuracy. Second, unlike fixed mask-based cropping strategies, ExpSlicer dynamically expands around predicted target regions, ensuring complete target coverage while preserving contextual information. To handle varying proposal counts, we set the proposal batch size to the average proposal number plus one, ensuring that over 50\% of images fully contribute to training.

\subsection{Attention Prior-Guided Knowledge Distillation (APKD)}

\begin{wrapfigure}[15]{r}{0.55\columnwidth}
% \vspace{-2em}
\centering
\includegraphics[width=\linewidth]{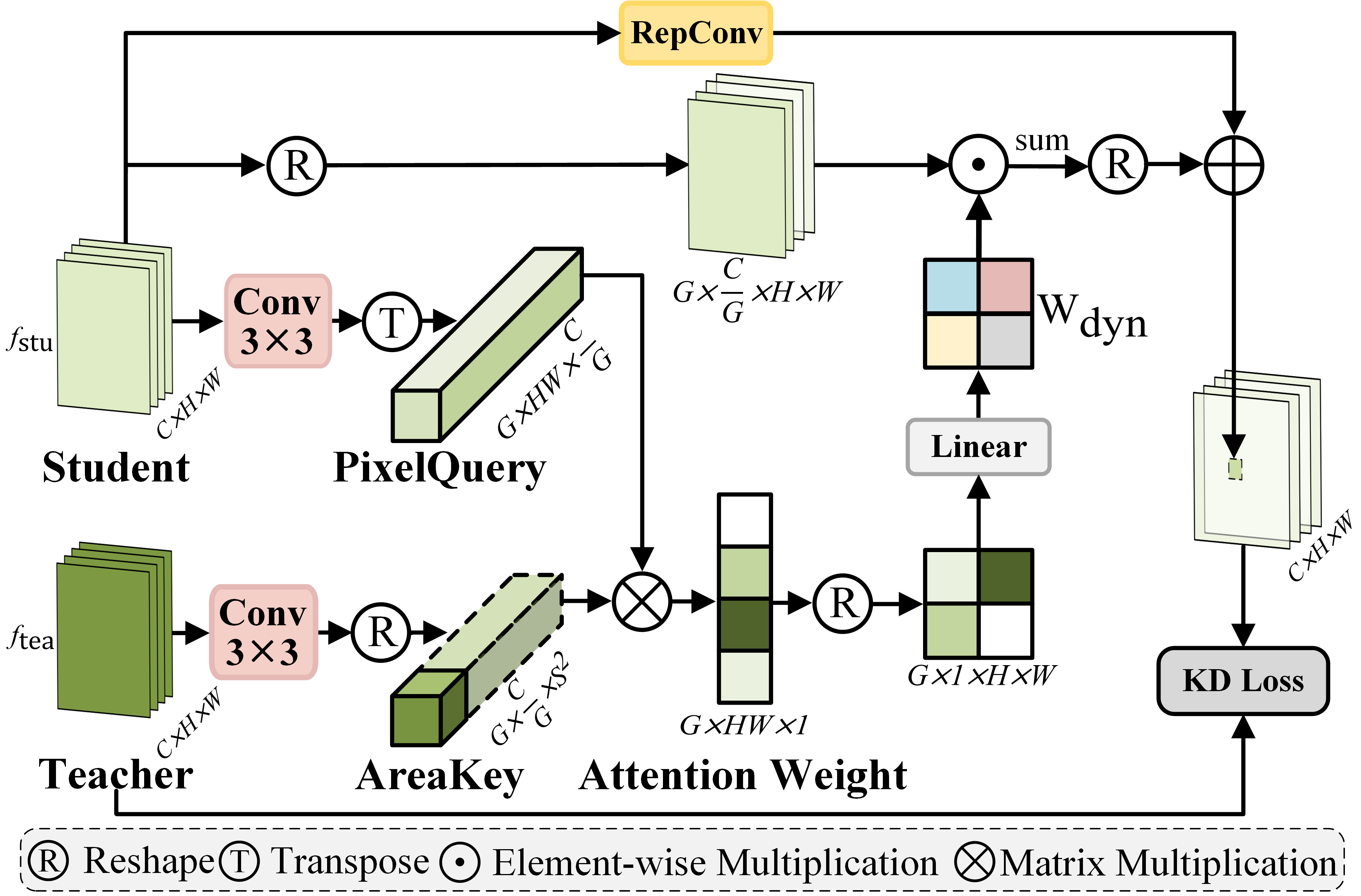} 
\caption{Overview of our APKD. }
\setlength{\abovecaptionskip}{2pt}
\label{knowledge_distillation} %\vspace{-1em}
\end{wrapfigure}
In this work, we customize a lightweight detector in the fine stage,  to balance accuracy and efficiency by incorporating channel-adaptive scaling and fewer detection heads. To further transfer the teacher’s attention priors over critical target regions, we propose an APKD, as shown in Fig. \ref{knowledge_distillation}. Mainstream conventional attention distillation methods typically flatten feature maps into 1D tokens, which weakens the preservation of spatial locality. To address this limitation, APKD performs teacher-student attention directly over multi-dimensional spatial features, explicitly maintaining the fine-grained spatial structure of IRSTs.  We adopt YOLOv12 \cite{2025NIPSTianYOLOv12} as the teacher model.

 % Conventional KD passively minimizes feature discrepancy, which is suboptimal for tiny targets where spatial misalignment dominates. We instead construct an explicit teacher-student cross-attention mechanism, enabling spatially-aware knowledge transfer.
 
%Instead of applying distillation at the output logits~\cite{2015arXivGeoffreyKD}, our method injects the teacher features directly into the spatial interaction process, empowering the student to dynamically adjust its perception of critical regions.
Given the student feature map $f_{\text{stu}} \in \mathbb{R}^{B \times C \times H \times W}$, we apply a $3 \times 3$ convolution to divide it into 
$G$ groups of lower-dimensional channels and obtain PixelQuery $\in \mathbb{R}^{B \times G \times \frac{C}{G} \times HW}$: $\text{PixelQuery} = \text{Conv}_{3 \times 3}(f_{\text{stu}}).$ This operation encodes local spatial details and fine-grained cues from the student. Meanwhile, the teacher's feature map $f_{\text{tea}}$ is processed similarly generating $S \times S$ AreaKey: $\text{AreaKey} =\text{Reshape} \left(\text{Conv}_{3 \times 3} (f_{\text{tea}}) \right)\in \mathbb{R}^{B \times G \times 
\frac{C}{G} \times S^2}$, where $S = \sqrt{H \cdot W}$.

Next, we compute their attention weight, denoted as $A$,  to global average representations of the feature map:
${A} = \text{Softmax}(\text{PixelQuery}^\top\cdot\text{AreaKey}),$ where $A$ encodes the spatial similarity between input tokens and globally pooled contextual representations from the teacher model. Subsequently, the affinity map is projected into convolution-like attention weights:
${W}_{\text{dyn}} = \text{Linear}(\text{Reshape}({A})).$

To further improve real-time performance and detection accuracy, we introduce three customized modifications. First, the channel dimensions of the neck network are adaptively aligned with the RBCN outputs to maintain feature consistency across stages and improve computational efficiency. Second, the original detection head is replaced with the stronger YOLOX head \cite{2021arxivYOLOXGe}, where sequential 4$\times$ and 2$\times$ channel compression is adopted instead of YOLO11’s uniform 4$\times$ compression, reducing computational cost while improving detection performance. Third, for IRSTD, the number of detection heads is reduced from three to two, with the two shallowest feature maps used as inputs. More details are provided in supplementary material.

%To further improve real-time performance and detection accuracy, three customized modifications are introduced: (1) the neck network channel dimensions are adaptively aligned with the RBCN outputs to maintain feature consistency across stages and improve computational efficiency; (2) the detection head is replaced with a much stronger YOLOX head \cite{2021arxivYOLOXGe}, adopting sequential 4$\times$ and 2$\times$ channel compression instead of YOLO11’s uniform 4$\times$ compression, achieving lower computational cost with better performance; (3) for IRSTD, the number of detection heads is reduced from three to two, using the two shallowest feature maps as input. More details are provided in supplementary material.

Unlike traditional knowledge distillation that passively aligns student features with those of the teacher, our approach actively modulates the student's perception via feature-level attention weight, making the student aware of semantically meaningful regions. This modulation is further enhanced through a context-guided residual gating generated by re-parameterize convolution (RepConv) \cite{ding2024CVPRUniRepLKNet}: % RepConv使用现成的方法增加引用！
$\text{output}=\text{Reshape}(\text{sum}(f_{\text{stu}} \odot W_{\text{dyn}}))+\text{RepConv}(f_{\text{stu}})$, where $\text{sum}(\cdot)$ denotes summation over the spatial dimensions of each channel. We define KD loss as $\mathcal{L}_{\text{KD}} = \sum_{i=1}^{L} w_i \cdot \mathcal{D}_{\cos}(F_s^i, F_t^i)$, 
where $F_s^i$ and $F_t^i$ denote the student and teacher features at the $i$-th level, respectively, and $w_i$ is the weight corresponding to that level, set to $\frac{1}{i}$.  $\mathcal{D}_{\cos}(\cdot)$ represents the cosine similarity.

% $\text{sum}(\cdot)$ denotes summation over each element after element-wise multiplicationzheli

 % By treating student features as queries and teacher features as keys, our APKD mechanism guides the student to focus on semantically critical regions, enabling the student to actively focus on and align with the crucial areas that teacher is concerned about. This teacher-student cross-attention is especially beneficial for lightweight detectors, enhancing target awareness while suppressing background noise without increasing inference cost.

Conventional cross-attention distillation methods typically flatten feature maps into 1D tokens, which weakens the preservation of spatial locality. To address this limitation, APKD performs attention directly over multi-dimensional spatial features, explicitly maintaining the fine-grained spatial structure of IRSTs. By using student features as queries and teacher features as keys, APKD guides the student to actively focus on teacher-highlighted critical regions. This teacher-student cross-attention is especially effective for lightweight detectors, improving target awareness and background suppression without increasing inference cost.

\section{Experiment}
\subsection{Datasets and Evaluation Metrics}
\noindent\textbf{Datasets.} We validate our method's effectiveness on three real IRST datasets, including UAV, Car, and IRSTD-1k \cite{2022CVPRZhangISNet}, under complex backgrounds. Our custom-built UAV and Car datasets comprise 98,420 and 18,230 infrared images, respectively, all with a spatial resolution of $640\times512$ pixels. Our datasets are divided into training, validation, and test sets in a ratio of 7:2:1. More details of three datasets (\textit{e.g.}, data source, target scale distribution, and background characteristics) are provided in the supplementary material.

\noindent\textbf{Evaluation Metrics.} To evaluate object-level detection performance, we adopt precision (P(\%)), recall (R(\%)), and $\text{AP}_{50}(\%)$ \cite{2023TIIFang}. For pixel-level evaluation, we utilize Pd(\%), Fa($10^{-6}$), and nIoU(\%) \cite{2022CVPRZhangISNet}. To evaluate the computational cost, we compute the floating-point operations (FLOPs, in G), the number of parameters (Params, in M), and the frames per second (FPS) with a batch size of 1 for a consistent comparison. We evaluate existing infrared segmentation methods by converting their predicted masks into detection boxes via the minimum bounding rectangles to calculate P, R, and AP \cite{2025AAAIZhangMOCID, 2024TGRSyangEFLNet, 2022CVPRWJeffriBox}. Furthermore, our method concentrates on coarsely recognizing the target-containing region and locating them in the fine stage, two specific metrics are employed for the ablation study,  Coarse Recognition Precision (CRP) and Coarse Recognition Recall (CRR):%\vspace{-0.6em}
%{{\scriptsize
\begin{align}
\text{CRP} = \frac{1}{M}\sum\limits_{j=1}^{M} \mathds{1} \left\{ \max\limits_{i=1,...,I} \left( \frac{{|\text{region}_{k\times k}^j }\cap {\text{box}^i|}}{|\text{region}_{k\times k}^j|} \right) \geq \tau \right\},\label{eqno16}\\
\text{CRR} = \frac{1}{I}\sum\limits_{i=1}^{I} \mathds{1} \left\{ \max\limits_{j=1,...,M} \left( \frac{|\text{region}_{k\times k}^j \cap \text{box}^i|}{|\text{box}^i|} \right) \geq \tau \right\} \label{eqno17}.
\end{align} %}}
Here, $M$ and $I$ denote the total number of predicted patches and GT boxes,  respectively. The selected regions after applying ExpSlicer and GT boxes are denoted as $\text{region}_{k \times k}$ and box. The thresholds $\tau$ are empirically set to $0.9$ in our experiments to ensure high spatial alignment between patches and targets. These metrics effectively evaluate the coarse stage: a high CRP indicates accurate region selection, while a high CRR reflects comprehensive coverage of true target regions, preserving essential contextual information.

\subsection{Implementation Details}
 We train our ECFNet with Adam  using a learning rate of 0.01. Both coarse-stage and fine-stage training lasts for 100 epochs, with a weight decay of $10^{-1}$ and a batch size of 12. Additionally, we employ dictinct object-level and pixel-level heads to evaluate our framework on different tasks (More details are provided in the supplementary material). We use random cropping, horizontal flipping, and brightness adjustment for augmentation. Following standard KD protocols, the teacher is trained with the same data, augmentations, and input resolution as the student. During distillation, the teacher remains frozen and only provide supervision signals. Our experiments are conducted on an NVIDIA RTX 4090 with CUDA 12.4 and PyTorch 2.5.  We select YOLOv12-L~\cite{2025NIPSTianYOLOv12}, YOLO11-L~\cite{2024GlennYOLO11} as the convolution-based detection methods, D-FINE-L~\cite{2025ICLRDFINE}, DQ-DETR~\cite{2024ECCVHuangDQ-DETR} and DINO~\cite{2023ICLRZhangDINO} are the representative Transformer-based detection methods, ESOD-L ~\cite{2025TIPESODLiu} and QueryDet \cite{2022CVPRYangQueryDet} are two-stage detection methods, EFLNet \cite{2024TGRSyangEFLNet}, DAGNet \cite{2023TIIFang}, PConv (MSHNet) \cite{2025AAAIYangPinwheel}, GSFANet \cite{2025TGRSDengGSFANet}, MSHNet \cite{2024CVPRLiuMSHNet}, LESPS \cite{2023CVPRYingLESPS}, UIU-Net \cite{2023TIPXUIUNet}, ISNet \cite{2022CVPRZhangISNet}, DNANet \cite{2023TIPBoDNANet} are designed for IRSTD.

\subsection{Quantitative Results}

As shown in ~\cref{exp_uav}, the object-level detection model D-FINE-L achieves relatively high P and R, yet it has a high computational cost. In contrast, the two-stage detection method ESOD-L reduces computational overhead, but it has lower P and R.  Our ECFNet,  while maintaining on par with other SOTA methods in terms of P, achieves the highest values for  R and AP$_{50}$. Compared with methods in ~\cref{exp_uav}, our model achieves an approximately {60\%} reduction in computational cost under comparable detection performance, and yields {+9.68} and {+8.58} improvements in P and R, under similar computational complexity. For pixel-level detection, ECFNet gains a {+2.30} and {+1.92} improvements in Pd and nIoU, compared to the second-best method.

% \begin{figure}[!t]
%     \centering
%     \includegraphics[width=2in,keepaspectratio]{PR_curve.png} \vspace{-1em}
%     \caption{P-R curves of our ECFNet and other SOTA methods on the UAV dataset. The area values under the curves are placed after the method names.}
%     \label{PR_curve}
% \end{figure}

\begin{table*}[!t]
    \centering
\fontsize{6}{7.8}\selectfont  %font size and line height
    \setlength{\tabcolsep}{0.5pt} %列间距
    \caption{Performance comparison against SOTA detectors on real infrared UAV, Car and IRSTD-1k datasets. The best and second-best results are highlighted in \textbf{Bold} and \underline{underline} within each category (Object-Level and Pixel-Level) respectively.} %\vspace{-1em}
    \begin{tabular}{cc|ccc|ccc|ccc|ccc}
        \hline
        \multirow{2}{*}{\raisebox{-0ex}{Object-Level}} & \multirow{2}{*}{\raisebox{-0ex}{Pub'Year}} & \multicolumn{3}{c|}{UAV} & \multicolumn{3}{c|}{Car} & \multicolumn{3}{c|}{IRSTD-1k} & \multirow{2}{*}{\raisebox{-0ex}{FLOPs$\downarrow$}} & \multirow{2}{*}{\raisebox{-0ex}{Params$\downarrow$}} & \multirow{2}{*}{\raisebox{-0ex}{FPS$\uparrow$}}  \\
        \cline{3-11}  %\cmidrule(lr){3-11}
        % \cline{4-6} \cline{7-9} \cline{10-12}
         & & P$\uparrow$ & R$\uparrow$ & AP$_{50}$$\uparrow$ & P$\uparrow$ & R$\uparrow$ & AP$_{50}$$\uparrow$ & P$\uparrow$ & R$\uparrow$ & AP$_{50}$$\uparrow$ &  &  &  \\
        \hline %\midrule
         ESOD-L &TIP'25     & 86.28 &  83.05 & 85.63 & 94.75 & \underline{91.73} & 93.32 & 78.59& 76.37 & 78.33 & \underline{38.1} & 72.8 & \underline{86}  \\
         YOLOv12-L &NeurIPS'25 & \underline{95.92} & 87.91 & 93.12 & \underline{96.09} & 82.87 & 91.94 & \underline{87.26} & 72.78 & 81.61 & 88.9 & 26.4 & 76 \\
         D-FINE-L &ICLR'25     & 95.27 & \underline{91.32} & 93.18 & 94.38 & 91.63 & \underline{93.68} & 85.67 & 81.56 & 83.53 & 90.7 & 30.6 & 40 \\
         YOLO11-L &2024     & 95.53 & 89.85 & \underline{95.39} & 96.04 & 85.35 & 92.33 & 84.53 & 77.12 & 82.19  & 86.6 & \underline{25.3} & 78 \\
         DQ-DETR &ECCV'24    & 89.84 & 78.23 & 84.22 & 88.83 & 74.75 & 83.15 & 77.56 & 75.28 & 76.93 & 578.5 & 58.7 & 18 \\
         DINO &ICLR'23     & 94.36 & 83.28 & 91.27 & 95.62 & 87.13 & 92.46 & 79.66 & 72.47 & 74.28 & 398.6 & 47.0 & 26 \\
         QueryDet  &CVPR'22    & 64.59 & 77.16 & 64.16 & 75.81 & 80.28 & 76.37 & 61.92 & 60.93 & 61.69 & 888.4 & 35.3 & 9 \\
         EFLNet &TGRS'24   & 93.28 & 86.89 & 91.92 & 93.18 & 90.53 & 92.92 & 87.03 & \underline{81.70} & \underline{84.32}   & 102.2    & 38.3 & 69\\
         DAGNet &TII'23     & 88.72 & 80.34 & 85.31 & 83.59 & 86.43 & 84.86 & 84.59 & 78.82 & 81.23 & 170.7 & 150.5 & 40 \\
         \textbf{Ours} & - & \textbf{95.96} & \textbf{91.63} & \textbf{95.87} & \textbf{96.55} & \textbf{92.16} & \textbf{96.06} & \textbf{90.18} & \textbf{84.71} & \textbf{89.23} & \textbf{31.7}  & \textbf{22.7} & \textbf{90} \\
        \hline %\cline{1-11} %\cmidrule{1-11}
        Pixel-Level &  Pub'Year & Pd$\uparrow$ & Fa$\downarrow$ & nIoU$\uparrow$ & Pd$\uparrow$ & Fa$\downarrow$ & nIoU$\uparrow$ & Pd$\uparrow$ & Fa$\downarrow$ & nIoU$\uparrow$ &  FLOPs$\downarrow$ & Params$\downarrow$ & FPS$\uparrow$ \\
        \hline  %\cline{1-11} %\cmidrule{1-11}
         \makecell*[c]{PConv\\ (MSHNet)} & AAAI'25 & 81.26 & \underline{12.11} & \underline{54.23} & \underline{83.26} & 14.66 & \underline{55.98} & 92.20 & \underline{10.70} & \underline{67.93} & 47.1  & 15.6 & 72 \\
         GSFANet & TGRS'25 & \underline{82.73} & 14.81 & 51.96 & 81.86 &  19.25 & 48.87  & 92.92 & 20.17 & 60.52 & 63.6  & \underline{8.8} & 62 \\
         MSHNet &CVPR'24 & 77.22  &17.07 & 45.65 & 82.74 & \underline{11.74} & 51.67 & \underline{93.88} & 15.03 & 67.16 & \underline{38.2}  & 15.5 & \underline{78} \\
         LESPS   &CVPR'23  & 75.63 & 23.77 & 35.68 & 70.16 & 23.55 & 44.57 & 89.85 & 14.22 & 53.19 & 88.5  & \textbf{4.7} & 71 \\
         UIU-Net &TIP'23    & 79.75 & 18.54 & 40.37 & 76.54 & 18.82 & 46.51 & 92.93 & 26.87 & 61.11 & 140.6 & 54.5 & 51 \\
         ISNet  &CVPR'22   & 80.66 & 19.22 & 41.58 & 80.37 & 16.99 & 49.85 & 92.33  & 18.57 & 64.23 & 185.4   & 9.7 & 63 \\
         DNANet &TIP'22   & 74.69 & 18.83 & 33.69  & 69.13 & 16.55 & 43.23 & 93.27 & 17.61 & 62.73 & 89.3 & \textbf{4.7} & 47 \\
         \textbf{Ours} & - & \textbf{83.56} & \textbf{11.69} & \textbf{56.11} & \textbf{85.45} & \textbf{10.52} & \textbf{58.21} & \textbf{94.91} & \textbf{9.94} & \textbf{69.67} & \textbf{34.3}  & 24.1 & \textbf{85} \\
        \hline
    \end{tabular}
    \label{exp_uav}
\end{table*}

% We observe that lower theoretical complexity does not necessarily yield higher FPS (e.g., D-FINE: 90.7 FLOPs, 40 FPS, UIU-Net: 51 FPS with 140.6 FLOPs). General detectors often incur higher latency due to stronger feature extraction and multi-scale fusion modules, whereas infrared detectors are typically optimized for hardware-friendly pixel-level processing, achieving higher runtime efficiency despite larger FLOPs.
We observe that reducing the theoretical computation of general detectors does not inherently yield higher FPS (e.g., D-FINE: 40 FPS with 90.7 FLOPs). In contrast, some infrared detectors sustain higher FPS despite larger computation (e.g., UIU-Net: 51 FPS with 140.6 FLOPs). This difference lies in: general box-based detectors, equipped with stronger feature extraction and multi-scale fusion modules, incur higher inference latency despite lower theoretical complexity. Conversely, infrared detectors are typically optimized for hardware-friendly and tuned for pixel-level enhancement on particular infrared datasets. Such designs enable higher runtime efficiency without sacrificing performance.

\begin{figure*}[!t]
    \centering
    \includegraphics[width=4.8 in,keepaspectratio]{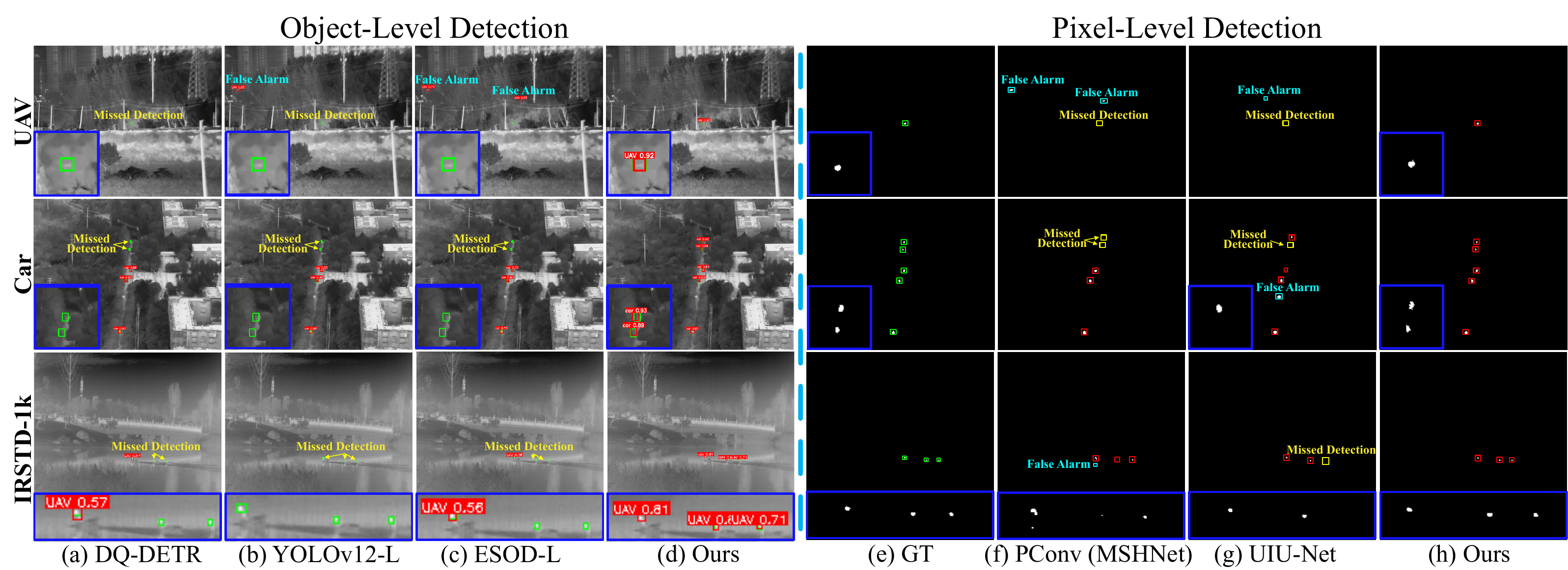} 
    \caption{Qualitative comparisons between our ECFNet and other SOTA methods on the UAV, Car and IRSTD-1k dataset. The green and red boxes denote the GT and the detection results. Closed-up views are shown in the left bottom corner and bottom.} %\vspace{-1.2em} 
    \label{qualitative_results}
\end{figure*}

\subsection{Qualitative Results}
From ~\cref{qualitative_results}, our method effectively detects IRSTs in three representative challenging backgrounds. Even in the presence of challenging background interference, our ECFNet accurately detects targets without false alarms. This is attributed to the DAT in the coarse stage for improved target–background discrimination, and the incorporation of APKD in the fine stage, which guides the student with attention priors from the teacher.  By effectively suppressing irrelevant background information before fine stage detection and retaining critical context information, our method ensures high precision and robustness for such scenes. The difficulty in suppressing background clutter in infrared images stems from their low texture and contrast, which limits the effectiveness of general-purpose detectors. Meanwhile, the sparse detection of two-stage frameworks like ESOD leads to insufficient contextual understanding, resulting in false alarms. Other methods tend to produce missed detections or false alarms.

\subsection{Ablation Study}
Extensive ablation experiments are conducted to validate the effectiveness of our method. More ablation studies and details of baselines in experiments (\cref{ablation_DAT_APKD_module}, \cref{tab:total_DAT_ablation}, \cref{tab:DAT_layer_ablation}, \cref{tab:apkd_ablation} and \cref{tab:Teacher_dependency}) are provided in the supplementary material.
% In this part, we conduct extensive ablation studies to validate the impact of ECFNet's main components. More ablation studies are provided in the supplementary material.

\begin{table*}[!t] %如果是单栏排版，请去掉星号改为 \begin{table}[!t]
    \centering  
    % ======= 第一个表格 =======
       \begin{minipage}{0.48\textwidth}
        \centering
        \fontsize{6}{9}\selectfont  % 统一设置字体大小
        \setlength{\tabcolsep}{0.8pt}  
        \caption{Ablation of the main components of our method. \textbf{Bold} denotes the best result.}
        %\vspace{-0.1em} % 根据需要调整间距
        \begin{tabular}{cc|ccc}
            \hline
            \multicolumn{2}{c|}{Model} & \multicolumn{3}{c}{Metrics} \\
            \hline %\midrule
            DAT & APKD & P$\uparrow$ & R$\uparrow$ & AP$_{50}$$\uparrow$  \\
           \hline %\midrule
            - & -  & 88.23 & 87.32 & 87.26   \\
            - & $\checkmark$ & 89.17\raisebox{-0.6ex}{\scriptsize(+0.94)} & 89.56\raisebox{-0.6ex}{\scriptsize(+2.24)} & 89.30\raisebox{-0.6ex}{\scriptsize(+2.10)}  \\
            $\checkmark$ & -  & 93.45\raisebox{-0.6ex}{\scriptsize(+5.22)} & 90.32\raisebox{-0.6ex}{\scriptsize(+3.00)} & 92.71\raisebox{-0.6ex}{\scriptsize(+5.45)}  \\
            $\checkmark$ & $\checkmark$ & \textbf{95.96}\raisebox{-0.6ex}{\scriptsize(+7.73)} & \textbf{91.63}\raisebox{-0.6ex}{\scriptsize(+4.31)} & \textbf{95.87}\raisebox{-0.6ex}{\scriptsize(+8.61)}  \\
            \hline
        \end{tabular}
        \label{ablation_DAT_APKD_module}
        \end{minipage}
    \hfill % 在两个表格之间添加弹性空白，使其均匀分布
    % ======= 第二个表格 =======
  \begin{minipage}{0.48\textwidth}
        \centering
        \fontsize{6}{9.4}\selectfont  %字体和行距
        \setlength{\tabcolsep}{0.8pt}
        \caption{Ablation study on the C2fP module in our RBCN. \textbf{Bold}: best, \underline{underline}: second best.}
        %\vspace{-0.17em} % 根据需要调整间距
        \begin{tabular}{c|cc|cc}
        \hline
        Module & CRP$\uparrow$ & CRR$\uparrow$ & FLOPs$\downarrow$  & Params$\downarrow$ \\
        \hline %\midrule
        Conv                       & 91.20 & 93.51 & 11.3  & 3.1  \\
        DWConv            & 90.06 & 92.57 & \textbf{1.8}  & \textbf{0.4} \\
        DCNv3                 & 92.17 & 94.42 & 9.8  & \underline{0.9} \\
        SwinTransformer     & \textbf{96.22} & \underline{95.41} & 34.5  & 4.1 \\
        C2fP                & \underline{94.13} & \textbf{ 95.69} & \underline{8.9}  & 2.6   \\
        \hline
        \end{tabular}
        \label{rbcn_ablation}
        %\vspace{-1em} % 根据需要调整间距
    \end{minipage}  
\end{table*}

% \begin{table}[!t]
% \centering
% \fontsize{8}{10}\selectfont  %font size and line height
% \setlength{\tabcolsep}{1.5pt}  
% \caption{Ablation of the main components of our method.}\vspace{-1em}
%     \begin{tabular}{cc|ccc}
%         \hline
%         \multicolumn{2}{c|}{Model} & \multicolumn{3}{c}{Metrics} \\
%         \hline
%     DAT & APKD & P $\uparrow$ & R $\uparrow$ & AP$_{50}$ $\uparrow$  \\
%     \hline
%     - & -  & 88.23 & 87.32 & 87.26   \\
%     - & $\checkmark$ & 89.17\raisebox{-0.6ex}{\scriptsize(+0.94)} & 89.56\raisebox{-0.6ex}{\scriptsize(+2.24)} & 89.30\raisebox{-0.6ex}{\scriptsize(+2.10)}  \\
%     $\checkmark$ & -  & 93.45\raisebox{-0.6ex}{\scriptsize(+5.22)} & 90.32\raisebox{-0.6ex}{\scriptsize(+3.00)} & 92.71\raisebox{-0.6ex}{\scriptsize(+5.45)}  \\
%     $\checkmark$ & $\checkmark$ & \textbf{95.96}\raisebox{-0.6ex}{\scriptsize(+7.73)} & \textbf{91.63}\raisebox{-0.6ex}{\scriptsize(+4.31)} & \textbf{95.87}\raisebox{-0.6ex}{\scriptsize(+8.61)}  \\
%     \hline
%     \end{tabular}
%     \label{ablation_DAT_APKD_module}\vspace{-1em}
% \end{table}

% \begin{figure}[!t]
% \centering
% \includegraphics[width=3in,keepaspectratio]{DAT_visual.pdf}  \vspace{-1em}
% \caption{Comparison of feature maps and results from the RBCN at different stages with and without DAT. Blue boxes denote the recognition result, and green boxes indicate the GT box.} 
% \label{DAT_visual}\vspace{-1em}
% \end{figure}

\begin{figure}[!t]
     \centering
     % 第一张子图 (a)
     \begin{subfigure}[b]{0.49\textwidth}
         \centering
         \includegraphics[width=\textwidth]{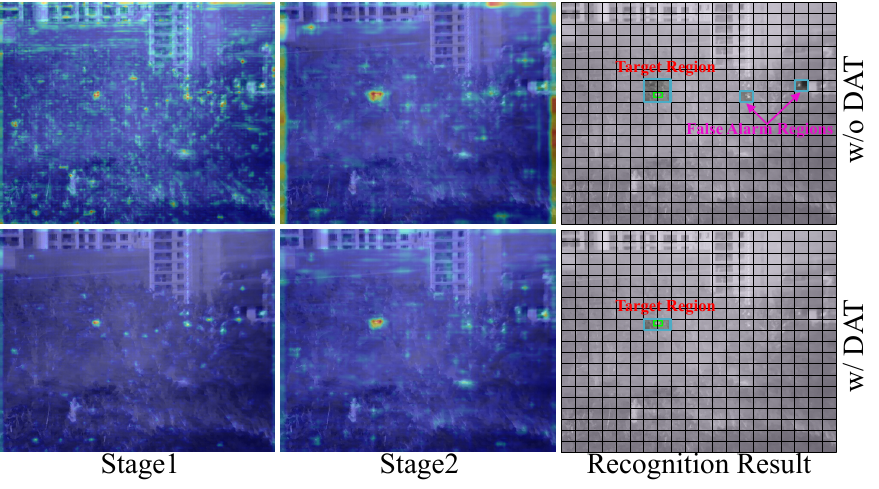}
         \caption{}
         \label{DAT_visual}
     \end{subfigure}
     \hspace{-0.5em} %较小的固定间距
     % 第二张子图 (b)
     \begin{subfigure}[b]{0.49\textwidth}
         \centering
         \includegraphics[width=\textwidth]{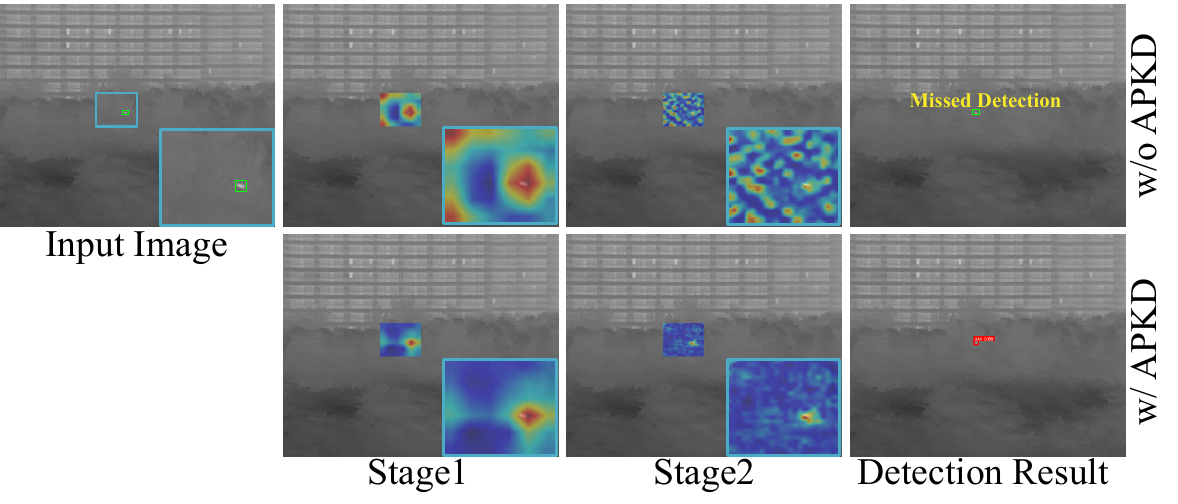}
         \caption{}
         \label{KD_visual}
     \end{subfigure}   
     \caption{(a) Comparison of feature maps and results from the RBCN at different stages with and without DAT. Blue boxes denote the recognition result, and green boxes indicate the GT box. (b) Comparison of feature maps and results of the lightweight fine detection model at different stages with and without APKD. Blue boxes denote the ExpSlicer's result. }
     \label{fig:total_visual_figure}
\end{figure}

\noindent\textbf{Impact of proposed main modules. } In ~\cref{ablation_DAT_APKD_module}, integrating either DAT or APKD consistently improves performance over our self-built baseline (row $1$). Incorporating APKD yields a $+2.10$ gain in {AP$_{50}$}, while introducing DAT brings a larger improvement of $+5.45$, highlighting the importance of high-quality coarse detection. When both modules are applied, performance peaks with increases of $+7.73$, $+4.31$, and $+8.61$ in P, R, and {AP$_{50}$}. These two components not only enhance their corresponding detection stages but also complement each other for optimal results. Specifically, DAT injects pseudo-sample noise to improve background–target discrimination, according to ~\cref{fig:total_visual_figure}(a), while APKD transfers spatial attention from a stronger teacher, guiding the lightweight student to focus on targets and suppress background interference as shown in ~\cref{fig:total_visual_figure}(b).

% \begin{figure}[!t]
% \centering
% \includegraphics[width=3.5in,keepaspectratio]{KD_visual.pdf} \vspace{-1em}
% \caption{Comparison of feature maps and results of the lightweight fine detection model at different stages with and without APKD. Blue boxes denote the ExpSlicer's result.} 
% \label{KD_visual}\vspace{-1em}
% \end{figure}

% \begin{table}[!t]
%     \centering
% \fontsize{8}{10}\selectfont  %font size and line height
%     \setlength{\tabcolsep}{1.4pt} %line space
%     \caption{Ablation study on the C2fP module in our RBCN.  \textbf{Bold} and \underline{underline} indicate the best and the second best results.}
% \begin{tabular}{c|cc|ccc}
% \hline
% Module & CRP $\uparrow$ & CRR $\uparrow$ & FLOPs (G) $\downarrow$  & Param (M) $\downarrow$ & FPS $\uparrow$\\
% \hline
% Conv                       & 91.20 & 93.51 & 11.34  & 3.10 & \underline{192} \\
% DWConv~\cite{2017CVPRCholletDWConv}             & 90.06 & 92.57 & \textbf{1.79}  & \textbf{0.37} & \textbf{200}\\
% DCNv3~\cite{2023CVPRWangDCNv3}                & 92.17 & 94.42 & 9.84  & \underline{0.85} & 162\\
% SwinTransformer \cite{2021ICCVLiuSwin}    & \textbf{96.22} & \underline{95.41} & 34.48  & 4.14 & 120\\
% C2fP                & \underline{94.13} & \textbf{ 95.69} & \underline{8.94}  & 2.57  & 140 \\
% \hline
% \end{tabular}
% \label{rbcn_ablation}
% \end{table}

\noindent\textbf{Impact of the main module in RBCN.}
In RBCN, the C2fP is the main feature extractor. As shown in~\cref{rbcn_ablation}, it achieves the highest CRR with competitive CRP and moderate computational cost. Although SwinTransformer~\cite{2021ICCVLiuSwin} attains slightly higher CRP, it incurs the highest FLOPs. Since the fine stage refines region proposals, the coarse stage can tolerate a few false alarms as long as it preserves high CRR. Unlike modules that improve accuracy through heavier computation, the proposed C2fP enhances directional awareness and spatial modeling. This design enriches feature representation with negligible overhead, achieving better accuracy–efficiency trade-offs than both heavy (DCNv3~\cite{2023CVPRWangDCNv3}, SwinTransformer) and lightweight (Conv, DWConv~\cite{2017CVPRCholletDWConv}) counterparts.

\begin{table}[!t]
    \centering
    \small
    \setlength{\tabcolsep}{1.5pt}
   \fontsize{8}{10}\selectfont  %字体大小和行距
    \caption{Ablation study on patch resolution predicted by RBCN. CRP and CRR evaluate region proposal quality, while AP$_{50}$ reflects its impact on fine detection. \textbf{Bold} indicate the best results.}
    \begin{tabular}{c|ccc|cc}
    \hline
    Patches & CRP $\uparrow$ & CRR $\uparrow$ & AP$_{50}$ $\uparrow$ & FLOPs (G) \textdownarrow & Params (M) \textdownarrow \\
    \hline
    $10\times10$ & 93.42 & 95.57 & 93.84 & \textbf{26.4} & 15.9 \\
    $20\times20$ & \textbf{94.13} & \textbf{95.69} & \textbf{95.7} & 26.9 & 15.9 \\
    $40\times40$ & 82.64 & 79.43 & 81.62 & 28.5 & 15.9 \\
    \hline
    \end{tabular}
    \label{entire_image_patches_ablation}
\end{table}

\begin{table}[!t] %!t强制放顶端
    \centering    
    % --- 左侧独立表格 (Table X) ---
    \begin{minipage}{0.48\textwidth}
        \centering
        % 独立标题和编号
        \caption{Ablation study of DAT. \textbf{Bold} and \underline{underline} indicate the best and the second best results.} 
        \label{tab:total_DAT_ablation}
        \fontsize{8}{11}\selectfont  %表字体大小和行间距
        \setlength{\tabcolsep}{2pt}%表列间距
        %\vspace{-0.17em} % 调整表头文字和表行间距
        \begin{tabular}{c|cc}
            \hline
            Method & CRP$\uparrow$ & CRR$\uparrow$ \\
            \hline %\midrule
            -        & 89.38 & 93.45    \\
            +Loss    & \underline{92.80} & \underline{94.35}   \\
            +GT      & 92.62 & 94.11   \\
            +Noise   & \textbf{94.13} & \textbf{95.69}    \\
            \hline
        \end{tabular}
    \end{minipage}%
    \hfill % 在两个表格之间添加弹性空白，使其均匀分布
    %\hspace{-0.0001em} %较小的固定间距
    % --- 右侧独立表格 (Table Y) ---
    \begin{minipage}{0.48\textwidth}
        \centering
        % 独立标题和编号
        \caption{Ablation study on inserting DAT into different basic blocks of RBCN. \textbf{Bold}: best, \underline{underline}: second best.} 
        \label{tab:DAT_layer_ablation}
        \fontsize{8}{9}\selectfont  %表字体大小和行间距 
        \setlength\tabcolsep{1.7pt} %表列间距
        %\vspace{-0.17em} %调整表头文字和表行间距
        \begin{tabular}{ccc|cc|ccc|cc} 
            \hline
            \multicolumn{3}{c|}{Blocks} & \multicolumn{2}{c|}{Metrics} & \multicolumn{3}{c|}{Blocks} & \multicolumn{2}{c}{Metrics} \\ 
            \hline %\midrule
            1 & 2 & 3 & CRP$\uparrow$ & CRR$\uparrow$ & 1 & 2 & 3 & CRP$\uparrow$ & CRR$\uparrow$ \\ 
           \hline %\midrule
             - & - & -  & 89.38 & 93.45 & \checkmark & \checkmark & - & 92.62 & 93.57 \\
            \checkmark & - & - & 92.67 & 93.52 & \checkmark& - & \checkmark& 93.80 & \underline{94.72} \\
             - & \checkmark & - & 91.19 & 91.36 & - & \checkmark & \checkmark& \underline{93.83} & 94.06 \\
             - & - & \checkmark & 91.77 & 93.35 & \checkmark & \checkmark & \checkmark & \textbf{94.13} & \textbf{95.69} \\ 
             \hline
        \end{tabular}
   \end{minipage}  
\end{table}

% \noindent\textbf{Impact of patch resolution predicted by RBCN.} As shown in \cref{entire_image_patches_ablation}, we evaluate the model's robustness to varying patch resolutions by comparing three grid configurations. While the input resolution is kept constant, the coarse stage outputs feature maps at different spatial resolutions. The results demonstrate that our region recognition is inherently stable; for instance, the $10\times10$ setting yields high CRP and CRR values that are closely comparable to the $20\times20$ ($32 \times 32$ pixel size region every patch) configuration. However, dividing the image into larger patches ($10\times10$, $64\times64$ pixel size region every patch) inevitably introduces more background clutter into the subsequent lightweight detector, causing a slight drop in AP$_{50}$. Conversely, adopting a finer grid ($40\times40$, $16\times16$ pixel size every patch) restricts the contextual receptive field for each patch. This increase the number of predictions (400 vs. 1600) and significantly elevates the task complexity, leading to a substantial performance degradation in the coarse stage. Given the weak signature of IRSTs, lacking sufficient surrounding context increases the difficulty of distinguishing target-containing regions, leading to a performance dip and higher computational cost. Therefore, we adopt the $20\times20$ resolution as the optimal sweet spot, as it robustly balances essential semantic context retention with effective clutter suppression.
\noindent\textbf{Impact of patch resolution predicted by RBCN.}
As shown in \cref{entire_image_patches_ablation}, we compare different RBCN patch resolutions under the same input size. The $10\times10$ grid achieves CRP and CRR close to the $20\times20$ setting, showing the robustness of RBCN. However, larger patches introduce more background clutter to the fine detector and slightly reduce AP$_{50}$, while the finer $40\times40$ grid lacks sufficient context for weak IRSTs and increases prediction complexity. Thus, we adopt the $20\times20$ resolution as a balanced setting that preserves semantic context while suppressing clutter.

\begin{figure}[!t]
     \centering
     % 第一张子图 (a)
     \begin{subfigure}[b]{0.37\textwidth}
         \centering
         \includegraphics[width=\textwidth]{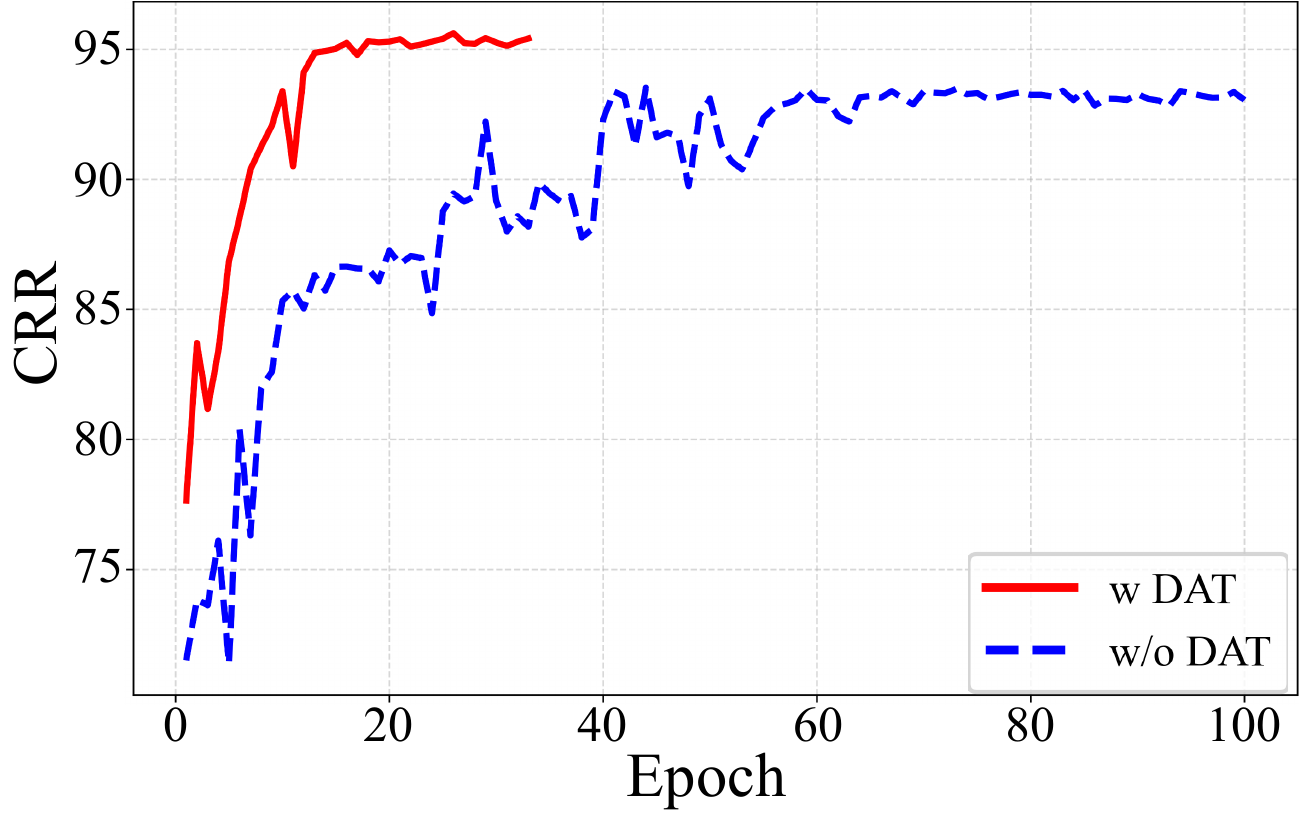}
         \caption{}
         \label{DAT_plot}
     \end{subfigure}
     % \hspace{-0.5em} %较小的固定间距
     % 第二张子图 (b)
     \begin{subfigure}[b]{0.58\textwidth}
         \centering
         \includegraphics[width=\textwidth]{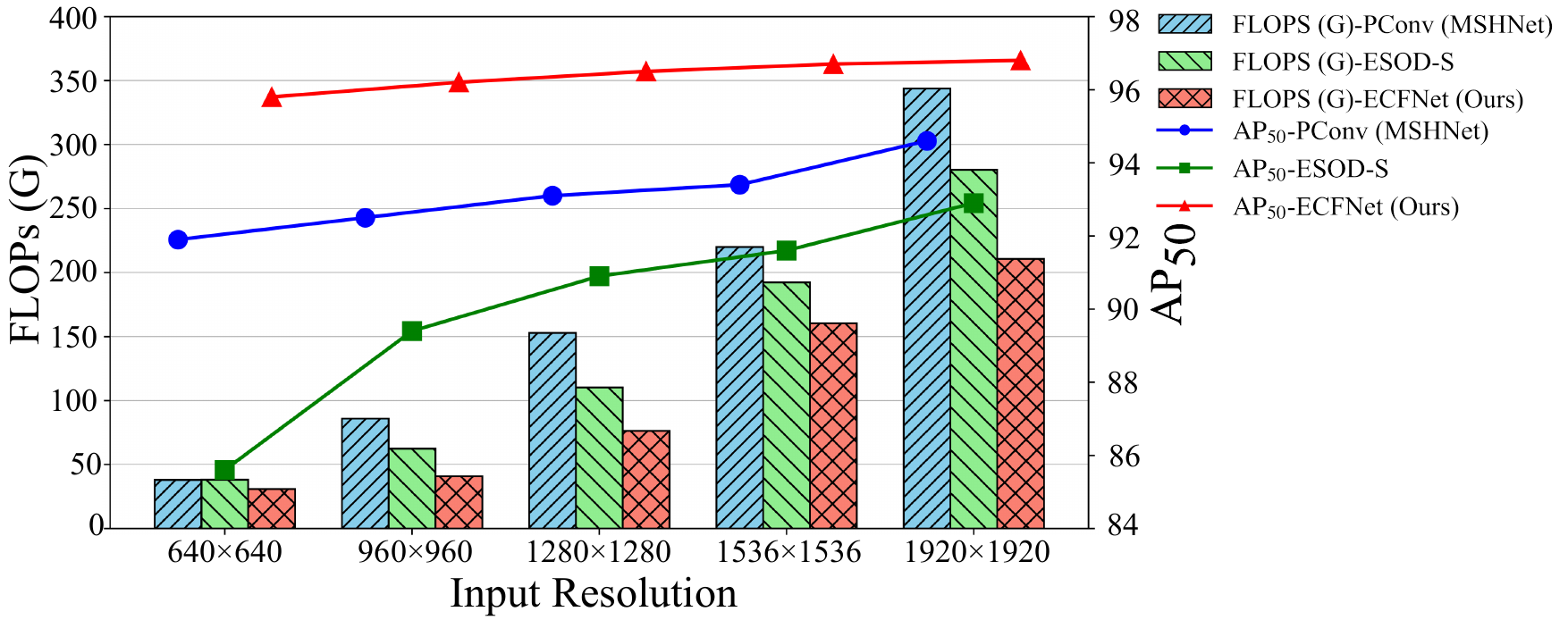} %\vspace{-1em}
        \caption{}  
        \label{High_resolution_exp}
     \end{subfigure}   
     \caption{(a) Convergence curves of the RBCN with and without the DAT module. (b) Performance comparison among PConv (MSHNet), ESOD-S, and our ECFNet. The average computation is reported only for cases with coarse-stage results.}
     \label{final_ablation}
\end{figure}

\noindent\textbf{Impact of DAT.} As shown in ~\cref{tab:total_DAT_ablation}, we evaluate the effectiveness of DAT under four configurations. Moreover, compared with directly computing the loss on the feature map, simply concatenating the GT mask even leads to a performance drop. This is because the direct concatenation of the GT mask weakens the model’s learning capability~\cite{2024TPAMILiDNDETR} and fails to optimize the backbone feature extraction effectively. In contrast, incorporating our DAT during training achieves the best overall performance. As illustrated in \cref{final_ablation}(a),  the inclusion of DAT significantly accelerates the training convergence speed and achieves a higher performance ceiling compared to the baseline. 

\noindent\textbf{Impact of DAT insertion locations.} As detailed in \cref{tab:DAT_layer_ablation}, although applying DAT to individual blocks provides consistent gains, deploying it globally across all RBCN blocks yields the best results.

\noindent\textbf{Impact of high-resolution images on complexity and accuracy.} In \cref{final_ablation}(b), we compare the FLOPs and AP of PConv (MSHNet), ESOD-S, and our ECFNet on  high-resolution images. It demonstrates that with the same input resolution, our method consistently reduces the FLOPs by up to about 40\% while achieving better performance in IRSTD.

% \begin{figure}[!t]
% \centering
% \includegraphics[width=3.2in,keepaspectratio]{High_resolution_exp.pdf} \vspace{-1em}
% \caption{Performance comparison among PConv (MSHNet), ESOD-S, and our ECFNet. The average computation is reported only for cases with coarse-stage results.}  
% \label{High_resolution_exp}\vspace{-1em}
% \end{figure}

\noindent\textbf{Impact of knowledge distillation strategy.} As shown in ~\cref{tab:apkd_ablation}, we conduct experiments using several knowledge distillation methods on our ECFNet. It can be seen that our APKD achieves the best results on all evaluated metrics, surpassing the second-best approach by up to {$4.72$} in AP$_{50}$ and {$1.31$} in R. Furthermore, as shown in \cref{tab:Teacher_dependency}, three distinct teacher models with up to a $7\times$ difference in complexity all yield performance gains exceeding $2.15$ in $\text{AP}_{50}$. This demonstrates APKD's robustness to teacher selection.

\begin{table}[!t]
    \centering    
    % --- 左侧独立表格 (Table X) ---
    \begin{minipage}{0.46\linewidth} % 使用 \linewidth 确保在单栏内按比例分配
        \centering
        \caption{Ablation study of APKD. \textbf{Bold} and \underline{underline} indicate the best and the second best results.}
        \label{tab:apkd_ablation}
        \fontsize{8}{8.5}\selectfont  %字体和行距
        \setlength{\tabcolsep}{1.5pt} %稍微压缩列宽，防止两表重叠
        %\vspace{-0.17em} %调整表头文字和表行间距
        \begin{tabular}{c|ccc}
             \hline
            Method & P$\uparrow$ & R$\uparrow$ & AP$_{50}$$\uparrow$ \\
            \hline %\midrule %空间不足, 请将\midrule改为\hline
            - & 93.45 & \underline{90.32} & 92.71 \\
            Plain~\cite{2015ICLRAdrianaFeatureKD} & 95.04 & 85.72 & \underline{91.15} \\
            DRKD~\cite{2023IJCAINiDRKD} & 95.17 & 88.51 & 90.13 \\
            CrossKD~\cite{2024CVPRWangCrossKD} & 93.82 & 84.14 & 87.70 \\
            DCSF~\cite{2025AAAIDaiDCSF} & \underline{95.82} & 86.73 & 89.29 \\
            APKD (Ours) & \textbf{95.96} & \textbf{91.63} & \textbf{95.87} \\
             \hline
        \end{tabular}
    \end{minipage}%
    \hfill % 弹性间距，自动把左右两表推向两端，中间留空
    % --- 右侧独立表格 (Table Y) ---
    \begin{minipage}{0.52\linewidth}
        \centering
        \caption{Ablation study of APKD’s teacher model. \textbf{Bold}: best, \underline{underline}: second best. $*$ denotes the default teacher in our APKD.} 
        % 注意：表内容看起来是 Teacher 消融，但保留了您的原标题文字
        \label{tab:Teacher_dependency}
        \fontsize{8}{8.5}\selectfont %字体和行距
        \setlength{\tabcolsep}{1.3pt}
        %\vspace{-0.17em} %调整表头文字和表行间距
        \begin{tabular}{c|cccc} 
            \hline
            Teacher & P$\uparrow$ & R$\uparrow$ & AP$_{50}$$\uparrow$ & FLOPs$\downarrow$ \\
            \hline %\midrule 
            - & 93.45 & 90.32 & 92.71 & - \\  
            PConv (MSHNet)& 94.82 & 90.63 & 93.72 & \textbf{12.4} \\ % 略微缩短了名字防止换行
            ESOD-L & 94.08 & 90.89 & 93.32 & 38.1 \\
            YOLO11-L & \underline{95.39} & \underline{91.27} & \underline{95.43} & \underline{86.6} \\ 
            ISNet & 94.31 & 90.28 & 92.54 & 185.4 \\
            YOLOv12-L$^*$ & \textbf{95.96} & \textbf{91.63} & \textbf{95.87} & 88.9 \\ 
            \hline
        \end{tabular}
    \end{minipage}    
\end{table}

\section{Conclusion}
In this paper, we propose ECFNet, an efficient coarse-to-fine infrared small target detection framework. In the coarse stage, we introduce RBCN to identify target region proposals, while DAT enhances target-background discrimination through a denoising task and  accelerate convergence. In the fine stage,  we tailor a lightweight detector to efficiently detect the target, and APKD guides it toward critical target regions via cross-attention, thereby enhancing discriminative feature representation for IRSTs. Experimental results demonstrate that ECFNet outperforms SOTA methods under complex backgrounds while maintaining high efficiency, making it suitable for embedded deployment.

% In this paper, we propose ECFNet, an efficient coarse-to-fine infrared small target detection framework. In the coarse stage, we introduce RBCN to identify target region proposals from complex backgrounds. In addition, DAT is leveraged to further enhance the RBCN’s ability to distinguish targets from cluttered backgrounds through a denoising task, while significantly accelerating model convergence and improving performance. In the fine stage, we tailor a lightweight detector to efficiently detect the targets and further exploit the APKD to guide the detector to focus on the critical target regions through cross-attention, thereby enhancing discriminative feature representation for IRSTs. Experimental results demonstrate that our ECFNet outperforms SOTA approaches on IRSTs with complex backgrounds and is promising for deployment on embedded platforms.

\section*{Acknowledgments}
The authors thank the Area Chair and reviewers for their valuable time and insightful comments. This work was supported by the Open Research Fund of the National Key Laboratory of Multispectral Information Intelligent Processing Technology under Grant 61421132301, and in part by the National Natural Science Foundation of China under Grants 62371203 and 62301228.

\bibliographystyle{splncs04}
\bibliography{main}

\begin{thebibliography}{10}
\providecommand{\url}[1]{\texttt{#1}}
\providecommand{\urlprefix}{URL }
\providecommand{\doi}[1]{https://doi.org/#1}

\bibitem{2024TGRSChen}
Chen, S., Ji, L., Zhu, S., Ye, M., Ren, H., Sang, Y.: Toward dense moving
  infrared small target detection: New datasets and baseline. IEEE Transactions
  on Geoscience and Remote Sensing  \textbf{62},  1--13 (2024)

\bibitem{2017CVPRCholletDWConv}
Chollet, F.: Xception: Deep learning with depthwise separable convolutions. In:
  Proceedings of the IEEE Conference on Computer Vision and Pattern Recognition
  (CVPR). pp. 1800--1807 (2017)

\bibitem{2025AAAIDaiDCSF}
Dai, T., Lin, Y., Guo, H., Wang, J., Zhu, Z.: {DCSF-KD}: Dynamic channel-wise
  spatial feature knowledge distillation for object detection. Proceedings of
  the AAAI Conference on Artificial Intelligence  \textbf{39}(3),  2627--2635
  (2025)

\bibitem{2025TGRSDengGSFANet}
Deng, C., Zhao, Z., Xu, X., Xia, Y., Li, J., Plaza, A.: {GSFANet}: Global
  spatial-frequency attention network for infrared small target detection. IEEE
  Transactions on Geoscience and Remote Sensing  \textbf{63},  1--17 (2025)

\bibitem{ding2024CVPRUniRepLKNet}
Ding, X., Zhang, Y., Ge, Y., Zhao, S., Song, L., Yue, X., Shan, Y.:
  Unireplknet: A universal perception large-kernel convnet for audio video
  point cloud time-series and image recognition. In: 2024 IEEE/CVF Conference
  on Computer Vision and Pattern Recognition (CVPR) (2024)

\bibitem{2022TIMFang}
Fang, H., Ding, L., Wang, L., Chang, Y., Yan, L., Han, J.: Infrared small {UAV}
  target detection based on depthwise separable residual dense network and
  multiscale feature fusion. IEEE Transactions on Instrumentation and
  Measurement  \textbf{71},  1--20 (2022)

\bibitem{2026AAAI_TDCNet}
Fang, H., Guo, S., Chen, Q., Chang, Y., Yan, L.: Spatio-temporal context
  learning with temporal difference convolution for moving infrared small
  target detection. In: Proceedings of the AAAI Conference on Artificial
  Intelligence. pp.~1--9 (2026)

\bibitem{2023ACMMMFang}
Fang, H., Liao, Z., Wang, L., Li, Q., Chang, Y., Yan, L., Wang, X.: {DANet}:
  Multi-scale {UAV} target detection with dynamic feature perception and
  scale-aware knowledge distillation. In: Proceedings of the 31st ACM
  International Conference on Multimedia (ACM MM). pp. 2121--2130 (2023)

\bibitem{2023TIIFang}
Fang, H., Liao, Z., Wang, X., Chang, Y., Yan, L.: Differentiated attention
  guided network over hierarchical and aggregated features for intelligent
  {UAV} surveillance. IEEE Transactions on Industrial Informatics
  \textbf{19}(9),  9909--9920 (2023)

\bibitem{2025CVPRUniCD}
Fang, H., Wang, X., Li, Z., Wang, L., Li, Q., Chang, Y., Yan, L.:
  Detection-friendly nonuniformity correction: A union framework for infrared
  {UAV} target detection. In: Proceedings of the IEEE/CVF Conference on
  Computer Vision and Pattern Recognition (CVPR). pp. 11898--11907 (2025)

\bibitem{2024ECCVHuangDQ-DETR}
Huang, Y.X., Liu, H.I., Shuai, H.H., Cheng, W.H.: {DQ-DETR}: {DETR} with
  dynamic query for tiny object detection. In: Proceedings of the European
  Conference on Computer Vision (ECCV). pp. 290--305 (2024)

\bibitem{2024GlennYOLO11}
Jocher, G., Qiu, J.: {Ultralytics YOLO11} (2024)

\bibitem{2022TIPBoDNANet}
Li, B., Xiao, C., Wang, L., Wang, Y., Lin, Z., Li, M., An, W., Guo, Y.: Dense
  nested attention network for infrared small target detection. IEEE
  Transactions on Image Processing  \textbf{32},  1745--1758 (2023)

\bibitem{2024TPAMILiDNDETR}
Li, F., Zhang, H., Liu, S., Guo, J., Ni, L.M., Zhang, L.: {DN-DETR}: Accelerate
  {DETR} training by introducing query denoising. IEEE Transactions on Pattern
  Analysis and Machine Intelligence  \textbf{46}(4),  2239--2251 (2024)

\bibitem{2025CVPRLi}
Li, T., Ye, M., Wu, T., Li, N., Li, S., Tang, S., Ji, L.: Pseudo visible
  feature fine-grained fusion for thermal object detection. In: 2025 IEEE/CVF
  Conference on Computer Vision and Pattern Recognition (CVPR). pp. 6710--6719
  (2025)

\bibitem{2025TIPESODLiu}
Liu, K., Fu, Z., Jin, S., Chen, Z., Zhou, F., Jiang, R., Chen, Y., Ye, J.:
  {ESOD}: Efficient small object detection on high-resolution images. IEEE
  Transactions on Image Processing  \textbf{34},  183--195 (2025)

\bibitem{2024CVPRLiuMSHNet}
Liu, Q., Liu, R., Zheng, B., Wang, H., Fu, Y.: Infrared small target detection
  with scale and location sensitivity. In: Proceedings of the IEEE/CVF
  Conference on Computer Vision and Pattern Recognition (CVPR). pp.
  17490--17499 (2024)

\bibitem{2021ICCVLiuSwin}
Liu, Z., Lin, Y., Cao, Y., Hu, H., Wei, Y., Zhang, Z., Lin, S., Guo, B.: {Swin
  Transformer}: Hierarchical vision transformer using shifted windows. In:
  Proceedings of the IEEE/CVF International Conference on Computer Vision
  (ICCV). pp. 9992--10002 (2021)

\bibitem{2025CVPRLouOverlock}
Lou, M., Yu, Y.: {OverLoCK}: An overview-first-look-closely-next convnet with
  context-mixing dynamic kernels. In: Proceedings of the IEEE/CVF Conference on
  Computer Vision and Pattern Recognition (CVPR). pp. 128--138 (2025)

\bibitem{2023CVPRWMeethalCascaded}
Meethal, A., Granger, E., Pedersoli, M.: Cascaded zoom-in detector for high
  resolution aerial images. In: 2023 IEEE/CVF Conference on Computer Vision and
  Pattern Recognition Workshops (CVPRW). pp. 2046--2055 (2023)

\bibitem{2022CVPRWJeffriBox}
Murrugarra-Llerena, J., Kirsten, L., Jung, C.R.: Can we trust bounding box
  annotations for object detection? In: Proceedings of the IEEE/CVF Conference
  on Computer Vision and Pattern Recognition Workshops (CVPRW). pp. 4812--4821
  (2022)

\bibitem{2023IJCAINiDRKD}
Ni, Z.L., Yang, F., Wen, S., Zhang, G.: Dual relation knowledge distillation
  for object detection. In: Elkind, E. (ed.) Proceedings of the Thirty-Second
  International Joint Conference on Artificial Intelligence (IJCAI). pp.
  1276--1284 (2023)

\bibitem{2025ICLRDFINE}
Peng, Y., Li, H., Wu, P., Zhang, Y., Sun, X., Wu, F.: {D-FINE}: Redefine
  regression task of {DETR}s as fine-grained distribution refinement. In:
  Proceedings of the International Conference on Learning Representations
  (ICLR). pp. 1--13 (2025)

\bibitem{2015NIPSRenFasterRCNN}
Ren, S., He, K., Girshick, R., Sun, J.: {Faster R-CNN}: Towards real-time
  object detection with region proposal networks. In: Advances in Neural
  Information Processing Systems (NeurIPS). vol.~28, pp. 91--99 (2015)

\bibitem{2015ICLRAdrianaFeatureKD}
Romero, A., Ballas, N., Kahou, S.E., Chassang, A., Gatta, C., Bengio, Y.:
  {FitNets}: Hints for thin deep nets. In: Proceedings of the International
  Conference on Learning Representations (ICLR). pp. 1--10 (2015)

\bibitem{2025ICCVAasishDM-EFS}
Sharma, A.: {DM-EFS}: Dynamically multiplexed expanded features set form for
  robust and efficient small object detection. In: Proceedings of the IEEE/CVF
  International Conference on Computer Vision (ICCV). pp. 24569--24579 (2025)

\bibitem{AAAI2025HS-FPNShi}
Shi, Z., Hu, J., Ren, J., Ye, H., Yuan, X., Ouyang, Y., He, J., Ji, B., Guo,
  J.: {HS-FPN}: High frequency and spatial perception {FPN} for tiny object
  detection. In: Proceedings of the AAAI Conference on Artificial Intelligence.
  vol.~39, pp. 6896--6904 (2025)

\bibitem{2025NIPSTianYOLOv12}
Tian, Y., Ye, Q., Doermann, D.: {YOLOv12}: Attention-centric real-time object
  detectors. In: Advances in Neural Information Processing Systems (NeurIPS).
  pp. 1--14 (2025)

\bibitem{2017NIPSVasAttention}
Vaswani, A., Shazeer, N., Parmar, N., Uszkoreit, J., Jones, L., Gomez, A.N.,
  Kaiser, {\L}., Polosukhin, I.: Attention is all you need. In: Advances in
  Neural Information Processing Systems (NeurIPS). pp. 6000--6010 (2017)

\bibitem{2024CVPRWangCrossKD}
Wang, J., Chen, Y., Zheng, Z., Li, X., Cheng, M.M., Hou, Q.: {CrossKD}:
  Cross-head knowledge distillation for object detection. In: Proceedings of
  the IEEE/CVF Conference on Computer Vision and Pattern Recognition (CVPR).
  pp. 16520--16530 (2024)

\bibitem{2022TGRSWangIAAN}
Wang, K., Du, S., Liu, C., Cao, Z.: {Interior} attention-aware network for
  infrared small target detection. IEEE Transactions on Geoscience and Remote
  Sensing  \textbf{60},  1--13 (2022)

\bibitem{2024ECCVShuoQueryDenoise}
Wang, S., Jia, F., Mao, W., Liu, Y., Zhao, Y., Chen, Z., Wang, T., Zhang, C.,
  Zhang, X., Zhao, F., Ricci, E., Roth, S., Russakovsky, O., Sattler, T.,
  Varol, G.: Stream query denoising for vectorized {HD}-map construction. In:
  Proceedings of the European Conference on Computer Vision (ECCV). pp.
  203--220 (2024)

\bibitem{2023CVPRWangDCNv3}
Wang, W., Dai, J., Chen, Z., Huang, Z., Li, Z., Zhu, X., Hu, X., Lu, T., Lu,
  L., Li, H., et~al.: Internimage: Exploring large-scale vision foundation
  models with deformable convolutions. In: Proceedings of the IEEE/CVF
  Conference on Computer Vision and Pattern Recognition (CVPR). pp.
  14408--14419 (2023)

\bibitem{2026AAAI_JFD3}
Wang, X., Fang, H., Li, Q., Wang, L., Chang, Y., Yan, L.: Blur-robust detection
  via feature restoration: An end-to-end framework for prior-guided infrared
  {UAV} target detection. In: Proceedings of the AAAI Conference on Artificial
  Intelligence. pp.~1--9 (2026)

\bibitem{2023TIPXUIUNet}
Wu, X., Hong, D., Chanussot, J.: {UIU-Net}: {U-Net} in {U-Net} for infrared
  small object detection. IEEE Transactions on Image Processing  \textbf{32},
  364--376 (2023)

\bibitem{2024TGRSyangEFLNet}
Yang, B., Zhang, X., Zhang, J., Luo, J., Zhou, M., Pi, Y.: {EFLNet}: Enhancing
  feature learning network for infrared small target detection. IEEE
  Transactions on Geoscience and Remote Sensing  \textbf{62},  1--11 (2024)

\bibitem{2022CVPRYangQueryDet}
Yang, C., Huang, Z., Wang, N.: {QueryDet}: Cascaded sparse query for
  accelerating high-resolution small object detection. In: Proceedings of the
  IEEE/CVF Conference on Computer Vision and Pattern Recognition (CVPR). pp.
  13658--13667 (2022)

\bibitem{2025AAAIYangPinwheel}
Yang, J., Liu, S., Wu, J., Su, X., Hai, N., Huang, X.: Pinwheel-shaped
  convolution and scale-based dynamic loss for infrared small target detection.
  In: Proceedings of the AAAI Conference on Artificial Intelligence. vol.~39,
  pp. 9202--9210 (2025)

\bibitem{2022TPAMIYangSCRDet}
Yang, X., Yan, J., Liao, W., Yang, X., Tang, J., He, T.: {SCRDet++}: Detecting
  small, cluttered and rotated objects via instance-level feature denoising and
  rotation loss smoothing. IEEE Transactions on Pattern Analysis and Machine
  Intelligence  \textbf{45}(2),  2384--2399 (2023)

\bibitem{2023CVPRYingLESPS}
Ying, X., Liu, L., Wang, Y., Li, R., Chen, N., Lin, Z., Sheng, W., Zhou, S.:
  Mapping degeneration meets label evolution: Learning infrared small target
  detection with single point supervision. Proceedings of the IEEE/CVF
  Conference on Computer Vision and Pattern Recognition (CVPR) pp. 15528--15538
  (2023)

\bibitem{2016ACMYuIoU}
Yu, J., Jiang, Y., Wang, Z., Cao, Z., Huang, T.: {UnitBox}: An advanced object
  detection network. In: Proceedings of the 24th ACM international conference
  on Multimedia (ACM MM). pp. 516--520 (2016)

\bibitem{2023ICCVXiangCFINet}
Yuan, X., Cheng, G., Yan, K., Zeng, Q., Han, J.: Small object detection via
  coarse-to-fine proposal generation and imitation learning. In: 2023 IEEE/CVF
  International Conference on Computer Vision (ICCV). pp. 6294--6304 (2023)

\bibitem{2023ICLRZhangDINO}
Zhang, H., Li, F., Liu, S., Zhang, L., Su, H., Zhu, J., Ni, L., Shum, H.Y.:
  {DINO:} {DETR} with improved denoising anchor boxes for end-to-end object
  detection. In: Proceedings of the International Conference on Learning
  Representations (ICLR). pp. 1--12 (2023)

\bibitem{2025AAAIZhangMOCID}
Zhang, M., Ouyang, Y., Gao, F., Guo, J., Zhang, Q., Zhang, J.: {MOCID}: Motion
  context and displacement information learning for moving infrared small
  target detection. Proceedings of the AAAI Conference on Artificial
  Intelligence  \textbf{39}(10),  10022--10030 (2025)

\bibitem{2022CVPRZhangISNet}
Zhang, M., Zhang, R., Yang, Y., Bai, H., Zhang, J., Guo, J.: {ISNet}: Shape
  matters for infrared small target detection. In: Proceedings of the IEEE/CVF
  Conference on Computer Vision and Pattern Recognition (CVPR). pp. 877--886
  (2022)

\bibitem{2024TGRSDTNet_Zhang}
Zhang, N., Liu, Y., Liu, H., Tian, T., Ma, J., Tian, J.: {DTNet}: A specialized
  dual-tuning network for infrared vehicle detection in aerial images. IEEE
  Transactions on Geoscience and Remote Sensing  \textbf{62},  1--15 (2024)

\end{thebibliography}

\end{document}